\documentclass[10pt,twocolumn,letterpaper]{article}

\usepackage[pagenumbers]{wacv} 

\usepackage[accsupp]{axessibility}

\usepackage{times}
\usepackage{latexsym}
\usepackage{booktabs}
\usepackage{graphicx}

\usepackage{array}
\usepackage{tabularx}

\usepackage[most]{tcolorbox}
\tcbuselibrary{listings}
\usepackage{multicol}

\usepackage{multirow}
\usepackage{booktabs}
\usepackage{tabularx}
\usepackage[table]{xcolor}

\newlength\savewidth\newcommand\shline{\noalign{\global\savewidth\arrayrulewidth
  \global\arrayrulewidth 1pt}\hline\noalign{\global\arrayrulewidth\savewidth}}
\newcommand{\tablestyle}[2]{\setlength{\tabcolsep}{#1}\renewcommand{\arraystretch}{#2}\centering\footnotesize}
\renewcommand{\paragraph}[1]{\vspace{1.25mm}\noindent\textbf{#1}}

\newcolumntype{x}[1]{>{\centering\arraybackslash}p{#1pt}}
\newcolumntype{y}[1]{>{\raggedright\arraybackslash}p{#1pt}}
\newcolumntype{z}[1]{>{\raggedleft\arraybackslash}p{#1pt}}

\usepackage[T1]{fontenc}

\definecolor{wacvblue}{rgb}{0.21,0.49,0.74}
\usepackage[pagebackref,breaklinks,colorlinks,allcolors=wacvblue]{hyperref}

\def\wacvPaperID{1796} 
\def\confName{WACV}
\def\confYear{2027}

\title{A Systematic Benchmark for Evaluating Multimodal Personalized Reasoning in Vision-Language Models}

\author{Dohyun Kim$^{1, *}$ \quad\ Junyong Lhim$^{2,*}$ \quad Kwanyong Park$^{1, \dagger}$\\
$^{1}$University of Seoul, Republic of Korea \qquad $^{2}$Sungkyunkwan University, Republic of Korea\\
{\tt\small $^{1}$\{kdoh04034, kwanyong.park\}@uos.ac.kr \qquad $^{2}$lhimjy1105@skku.edu}
}

\begin{document}
\maketitle
{\renewcommand{\thefootnote}{}
\footnotetext{$^{*}$Equal contribution. \quad $^{\dagger}$Corresponding author.}}
\begin{abstract}
People often make decisions by holistically reasoning over heterogeneous forms of personal information. In this paper, we study the relatively underexplored capability of multimodal personalized reasoning in multimodal large language models (MLLMs). To this end, we introduce \texttt{MPRBench}, a novel and comprehensive benchmark specifically designed to evaluate personalized reasoning across a wide range of tasks and real-world challenges. \texttt{MPRBench} consists of 12 sub-tasks and additionally supports interpretable analysis of representative error cases. Through extensive experiments, we find that current personalized MLLMs still struggle with personalized reasoning due to several key challenges, including inaccurate recognition of personalized concepts and sensitivity to irrelevant personal information. We hope that MPRBench will stimulate further research on personalized reasoning in MLLMs. Code and data are available at \url{https://github.com/kkdh15/MPRBench}
\end{abstract}
    
\section{Introduction}
\label{sec:intro}

In daily life, people frequently interact with personal concepts such as friends, belongings, and pets, relying on their own memories. These memories are inherently multimodal and play a central role in human behavior and decision-making. For instance, when planning what to do, we often reason holistically by considering the visual appearance of close friends, their habits, past interactions, and the sentiments we expect from them. In other words, humans naturally integrate current context with heterogeneous forms of personal memory, including visual impressions, textual knowledge, past experiences, preferences, and social relationships.

\begin{figure}[t]
  \centering
  \includegraphics[width=\linewidth]{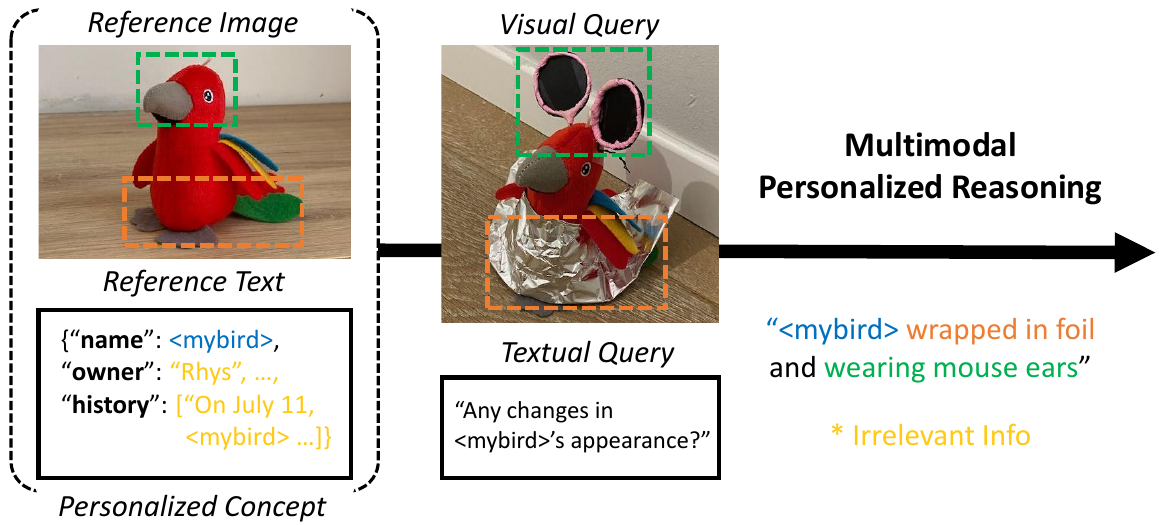}
  \vspace{-7mm}
  \caption{\textbf{Conceptual Overview and Problem Setup for Multimodal Personalized Reasoning.}}
  \label{fig:teaser}
  \vspace{-4mm}
\end{figure}

Given this nature of human reasoning, an ideal personalized AI assistant should be able to reason holistically over both the current situation and diverse forms of personal memory. Achieving this goal requires substantial advances in multimodal reasoning over personal information, particularly across heterogeneous modalities such as vision and language, which remains relatively underexplored in current research.

Although prior studies on personalized multimodal large language models (MLLMs)~\cite{nguyen2024yo,hao2025rap} have made significant progress, they address this challenge only partially. Existing approaches mainly focus on relatively simple tasks, such as recognizing the presence of personal concepts from visual inputs~\cite{nguyen2024yo} or reporting their basic status~\cite{an2024mc}. While current personalized models perform well on such benchmarks, these settings remain far from the complexity of real-world user requests. As a result, it is still unclear whether existing models can faithfully handle realistic scenarios and satisfy user needs, especially when deeper reasoning over multimodal personal information is required.

To bridge this gap, we introduce \texttt{MPRBench}, a novel and comprehensive benchmark for personalization, designed to evaluate the reasoning capabilities of MLLMs with multimodal personal information. In designing \texttt{MPRBench}, we consider emerging challenges in real-world personalized AI scenarios. In particular, we highlight a key challenge of multimodal personalized reasoning: \textit{the ability to reject irrelevant personal information during the reasoning process}. Although multimodal AI systems may have access to a wide range of personal information in both visual and textual forms, not every user request requires all of that information. Thus, a personalized MLLM should continually determine whether each piece of personal information is relevant to the current reasoning process; otherwise, it should rely on its internal generic knowledge instead.

Based on this insight, we divide our benchmark into three groups—visual-specific, text-specific, and multimodal reasoning—according to which modality is ideally required to solve the problem. Specifically, \texttt{MPRBench} consists of five visual-specific reasoning tasks, three text-specific reasoning tasks, and four multimodal reasoning tasks, resulting in a total of 12 sub-tasks. These design choices allow our benchmark to more faithfully reflect real-world personalized scenarios and to systematically analyze both the performance and failure cases of current models.

On \texttt{MPRBench}, we evaluate a wide range of personalized MLLMs spanning four major categories of prior work: (1) zero-shot, prompting-based personalization~\cite{seifi2025personalization}, (2) methods based on learning specialized personalized tokens~\cite{nguyen2024yo,an2024mc}, (3) retrieval-based models~\cite{hao2025rap,das2025training} and (4) memory-augmentation-based methods~\cite{hong2026tameing,nie2026personavlm}. Through extensive and systematic experiments, we obtain several key findings. 
First, robust recognition of personal concepts remains a fundamental bottleneck for personalized reasoning. Second, current methods struggle to reject irrelevant personal information across diverse scenarios. Third, reasoning that requires sequential and repeated use of personalized information remains highly challenging for existing models. Together, these findings highlight clear technical barriers that must be addressed.

Our contributions are summarized as follows: (1) We introduce \texttt{MPRBench}, the first comprehensive and interpretable benchmark for evaluating multimodal personalized reasoning.
(2) We provide a systematic analysis of current personalized MLLMs and identify key factors that limit their personalized reasoning capabilities.

\section{Related Works}

\noindent\textbf{Multimodal Large Language Models.}
Inspired by the rapid progress of large language models (LLMs)~\cite{touvron2023llama,bi2024deepseek}, extensive research efforts have been devoted to extending these advances to diverse modalities, including vision~\cite{liu2023visual} and audio~\cite{deshmukh2023pengi}. In particular, for the visual modality, multimodal large language models (MLLMs) have substantially advanced classical vision-language tasks such as visual question answering~\cite{schwenk2022okvqa,lee2026holisafe}, image captioning~\cite{krishna2017visual,saito2025captionsmiths} and visual grounding~\cite{yin2025rod,an2026talk,park2025learning,park2024weak}. More recently, the research focus has shifted toward building strong visual reasoning capabilities~\cite{johnson2017clevr}, where models are expected not only to perceive visual content but also to reason over it using commonsense and contextual understanding. To support this trend, many comprehensive multimodal benchmarks~\cite{li2023seed,yue2024mmmu,chen2024we,liu2024mmbench} have been proposed to cover a wide range of reasoning-oriented tasks. These benchmarks have played an important role in driving progress in visual reasoning and enabling the development of more capable MLLMs~\cite{xu2025llava, bai2025qwen3, guo2025deepseek,liu2025visual}.

Despite these advances, most prior work has focused on generic visual reasoning, in which reasoning is grounded primarily in common world knowledge. In contrast, our work investigates the relatively unexplored problem of personalized multimodal reasoning, where models must actively leverage multimodal personal information during the reasoning process.

\noindent\textbf{Personalization in MLLMs.}
Early works such as MyVLM~\cite{alaluf2024myvlm} and Yo’LLaVA
~\cite{nguyen2024yo} opened the research area of personalized MLLMs by focusing on recognizing personal concepts and understanding their states.
However, such approaches are limited in scalability because they require iterative training for each concept.
To overcome this limitation, subsequent studies have attempted to remove concept-wise training costs by introducing large-scale personalization tuning~\cite{hao2025rap,pi2025personalized,oh2026repic} or training-free retrieval-based methods~\cite{das2025training, seifi2025personalization}. 

More recently, several works have broadened the scope of personal information beyond visual identity, incorporating additional attributes such as user preferences~\cite{kim2026mmpb, kim2026pervl} or long-term interaction histories~\cite{hong2026tameing, nie2026personavlm}. Nevertheless, their target tasks still largely follow a recognition-and-retrieval paradigm, where models first identify a personalized concept and then retrieve or report the associated personal information.

In contrast, we systematically investigate diverse \textbf{personalized reasoning} tasks, in which each concept is represented not only by visual references but also by rich textual attributes. This setting requires models to go beyond concept recognition and attribute retrieval: they must identify relevant personal information and reason over it together with the current visual context and general commonsense knowledge across diverse reasoning scenarios.
\section{The MPRBenchmark}

\subsection{Overview of MPRBenchmark}
{Our goal is to assess the multimodal reasoning capabilities of personalized MLLMs in real-world scenarios. In particular, we aim to cover a wide range of reasoning tasks involving personal information and to provide insights for developing more robust personalized models. While existing benchmarks~\cite{nguyen2024yo,alaluf2024myvlm,das2025training,an2026unictokens,kim2026pervl} primarily evaluate elementary tasks, such as identifying personal entities, recognizing their states in visual contexts, or retrieving their associated attributes, MPRBench is designed to challenge higher-level reasoning tasks in which multimodal personal information must be used in diverse ways.}

\begin{figure*}[t]
    \centering
    \includegraphics[width=\linewidth]{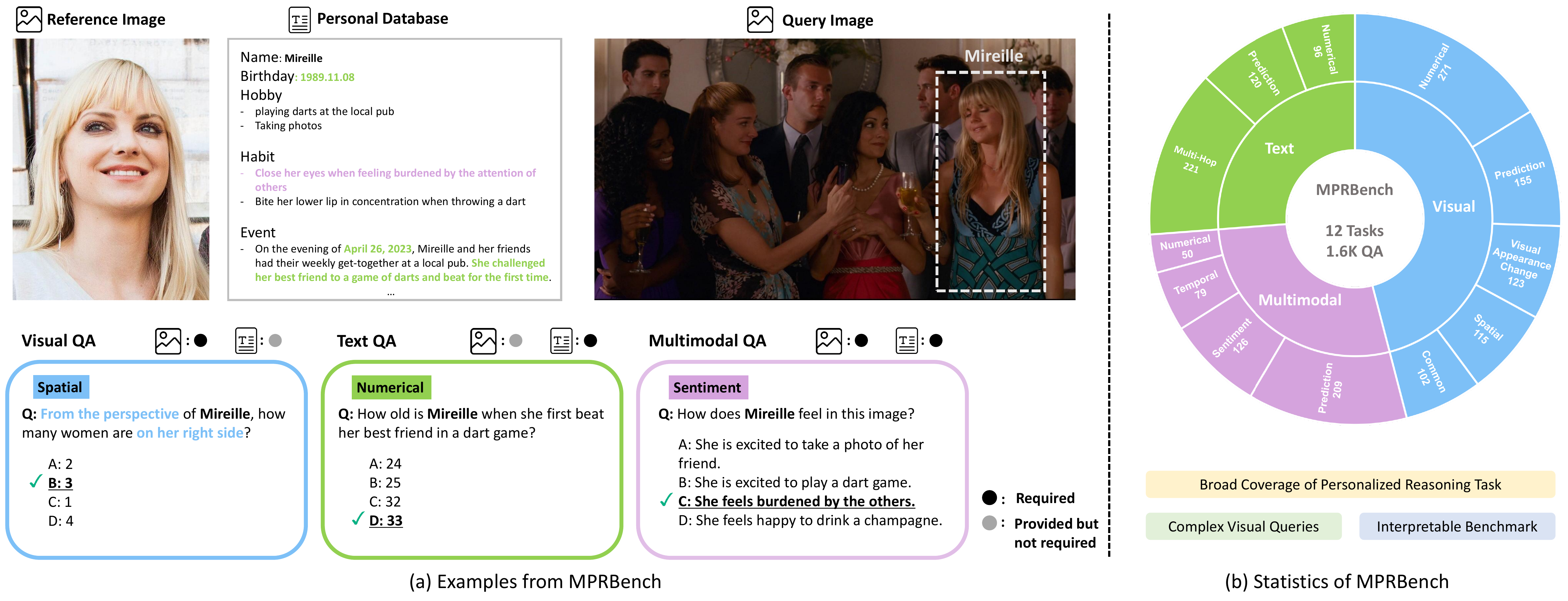}
    \caption{\textbf{Illustrative Examples and Statistics of MPRBench.}}
    \label{fig:example}
\end{figure*}

\noindent\textbf{Notions and Problem Setup.}
We define a user-specific entity as a \textit{personalized concept}, represented by a set of \textit{reference image(s)} and \textit{reference text}. The reference image(s) serve as a visual profile that captures the entity’s appearance and functions as a visual identifier. The reference text provides structured personal information in a JSON-like textual format, including attributes such as name, preferences (for persons), ownership (for objects), and relevant history.
Given this, the goal of each task is to generate an appropriate answer to a \textit{textual query} that mimics a user request, by jointly reasoning over both the \textit{visual query} representing the current circumstance and the known personalized concept. (See Fig.~\ref{fig:teaser})

In this setting, a personalized MLLM must effectively manage personalized concepts composed of reference images and text, and must seamlessly and selectively leverage the relevant information needed to answer the given query.

\subsection{Key Features of MPRBenchmark.}
\label{subsec:key}

\noindent\textbf{Broad coverage of personalized reasoning tasks.}
MPRBench is designed to enable holistic evaluation of personalized reasoning with multimodal personal information. Accordingly, it includes not only tasks that require joint reasoning over both visual and textual personal information, but also tasks for which only one modality is sufficient. Based on this taxonomy, we formulate realistic reasoning-oriented sub-tasks for each group. The resulting benchmark contains 12 sub-tasks in total, supporting evaluation across diverse scenarios and allowing fine-grained analysis of model weaknesses.

\noindent\textbf{Complex visual queries.}
To better reflect realistic reasoning scenarios, we intentionally construct image queries with complex scenes in which personalized concepts appear alongside other entities within cluttered backgrounds. This design differs substantially from existing benchmarks, which often rely on personalized-object-centric images~\cite{nguyen2024yo} (e.g., images containing only the target concept) or attribute-centric illustrations that omit the personalized concept itself~\cite{kim2026mmpb} (e.g., an image showing only a preferred concert). Although this choice significantly increases the burden of data construction, it enables more challenging high-level reasoning tasks and better represents the complexity of real-world environments.

\noindent\textbf{Interpretable benchmark design.}
We design MPRBench to support interpretable analysis of personalized models. Specifically, we formulate all tasks as multiple-choice question answering problems, where each option is designed to reflect a representative failure case. This design makes it easier to identify major failure modes and provides actionable insights into technical directions for future research.

\begin{figure*}[t]
    \centering
    \includegraphics[width=0.95\linewidth]{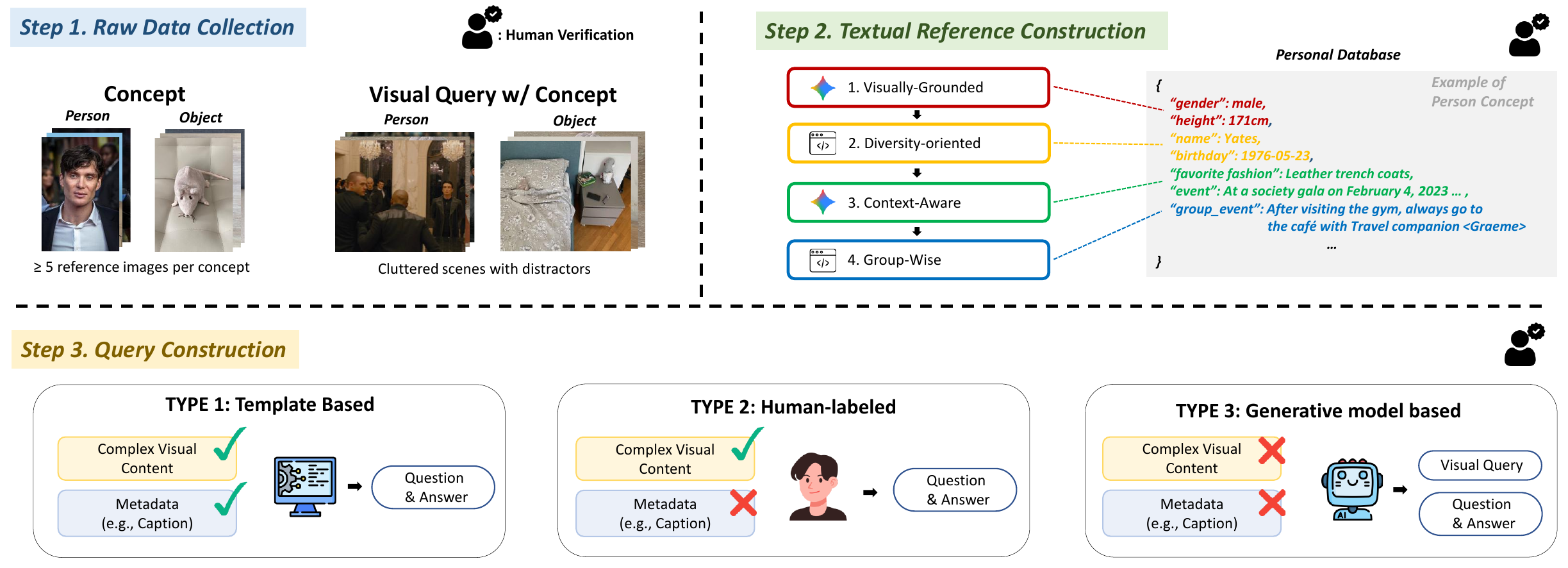}
    \caption{\textbf{Overview of the Data Construction Pipeline for MPRBench.}}
    \label{fig:data}
\end{figure*}

\subsection{Task Definition}
\label{sec:task_definition}

MPRBench comprises 12 sub-tasks, each designed to capture a distinct challenge in multimodal personalized reasoning. Below, we describe each task group and highlight the key reasoning demands they impose. Representative examples for selected sub-tasks are provided in Fig.~\ref{fig:example}, the complete set of examples is provided in the supplementary.

\noindent\textbf{Vision-specific reasoning.}
This group contains tasks for which only visual information—reference image(s) and the visual query—is necessary for successful reasoning. In these tasks, a personalized model must first accurately identify the personalized concept from the reference image(s), and then reason about the current visual scene in relation to that concept. We include personalized variants of several classical vision-oriented reasoning tasks~\cite{liu2024mmbench,chen2024we,ma20253dsrbench,zellers2019recognition}: (1) \textbf{predictive reasoning}, which requires forecasting the concept’s likely near-future action or situation; (2) \textbf{spatial reasoning}, which requires understanding spatial relations involving the concept; (3) \textbf{numerical reasoning}, which requires reasoning about counts or quantities related to the concept; and (4) \textbf{visual commonsense reasoning}, which requires inferring the rationale behind the visual state of the concept based on commonsense knowledge. In addition, we introduce (5) \textbf{visual appearance change reasoning}, a novel task motivated by realistic user needs such as determining whether a close friend has changed their hairstyle. This task requires explicit comparison between stored visual memory from the reference image(s) and the concept’s current appearance in the query image.

\noindent\textbf{Text-specific reasoning.}
In contrast, this group contains tasks for which textual information alone is sufficient, and visual information is unnecessary. These tasks require reasoning over textual descriptions of personalized concepts together with textual user queries. We include (1) \textbf{predictive reasoning}, which involves predicting plausible outcomes from personal preferences, habits, or prior experiences under a given condition, and (2) \textbf{numerical reasoning}, which involves arithmetic or logical reasoning over numerical personal attributes such as birthdays. We further introduce (3) \textbf{personalized multi-hop reasoning}, a novel task that requires sequentially combining multiple pieces of textual personal information. Inspired by generic multi-hop reasoning~\cite{ho2020constructing}, this task is designed to test whether models can repeatedly retrieve and integrate relevant personal attributes across multiple reasoning steps.

\noindent\textbf{Multimodal reasoning.}
This group consists of tasks in which both visual and textual personal information are jointly required. As such, it directly evaluates whether a model can align, retrieve, and integrate heterogeneous personal information across modalities. We include (1) \textbf{predictive reasoning}, which requires forecasting likely actions based on personal preferences or habits under a given visual context; (2) \textbf{sentimental reasoning}, which requires inferring the cause or rationale of visually expressed emotion in light of personal events or experiences; and (3) \textbf{numerical reasoning}, which requires counting or aggregating textual attributes associated with concepts present in the visual scene. Finally, we propose (4) \textbf{multimodal history reasoning}, a novel task in which the model must infer when a visual query was captured by matching visual evidence in the current image with historical information about the concept described in text.

We note that both the reference image(s) and reference text are provided to personalized MLLMs in order to reflect real-world conditions, where users often do not explicitly specify which information should be considered.

\subsection{Dataset Construction Process}
We design a semi-automatic dataset construction pipeline to reduce human labor while maintaining overall data quality. We focus here on the high-level construction process (See Fig.~\ref{fig:data}); more detailed descriptions are provided in the supplementary material.

\subsubsection{Raw Data Collection}
\label{sec:raw_data_collection}

\noindent\textbf{Concept collection.}
We begin by manually collecting reference images for personalized concepts. These images are sourced either from existing open datasets or from CC BY 4.0 licensed images collected from the internet. For person concepts, we use existing personalization-related datasets such as Yo’LLaVA~\cite{nguyen2024yo}, MC-LLaVA~\cite{an2024mc}, MMPB~\cite{kim2026mmpb} and CustomConcept101~\cite{kumari2023multi}, as well as images of sports players from SoccerAgent~\cite{rao2025multi} and publicly available CC BY 4.0 licensed images of celebrities. For object concepts, we repurpose ConConChi~\cite{rosasco2024concon} and RPC~\cite{wei2019rpc}, which were originally constructed for object retrieval and Retail Product Checkout tasks, respectively. To faithfully represent the visual appearance of each concept, we crop the original images to remove complex surrounding contexts. In addition, since some prior personalization methods~\cite{nguyen2024yo} require multiple reference images, we retain only concepts for which at least five reference images are available.

\noindent\textbf{Complex visual query collection.}
We also manually collect complex visual query images, with particular emphasis on scene complexity. Specifically, we prioritize images in which the target concept appears together with visually similar concepts (i.e., distractors) and under cluttered backgrounds. These conditions provide valuable opportunities to assess how concept identification difficulty affects downstream reasoning performance. Based on these criteria, we source person-related visual queries from VCR~\cite{zellers2019recognition} and SoccerAgent~\cite{rao2025multi}, and object-related queries from ConConChi and RPC. We make substantial efforts to ensure high quality during this stage. For example, we manually inspected approximately 4,000 images from VCR and selected only 120 images that satisfied our quality standards for visual queries.

\subsubsection{Textual Reference Construction}
\label{sec:person_construction}

Constructing high-quality textual references for each concept is a particularly important step, as it strongly affects the quality of subsequent data generation stages. We carefully design a set of core textual attributes: 16 attributes for person concepts, including name, age, favorites, habits, and life events, and 8 attributes for object concepts, including owner, acquisition process, and history. The full attribute list is provided in the appendix. These textual references should be simultaneously plausible, consistent with the visual appearance of the concept, diverse across the dataset, and well correlated with the visual query so as to support the construction of high-level reasoning tasks.

To achieve this, we structure the textual reference construction process into four steps. We describe the process for person concepts as an example. 

\noindent\textbf{Step 1. Visually-grounded Attributes.} We prompt Gemini-2.5-Pro~\cite{comanici2025gemini} to infer visually grounded attributes, such as age and gender, that are strongly correlated with the concept’s appearance in the reference images. 

\noindent\textbf{Step 2. Diversity-oriented Attributes.} To promote diversity, we randomly assign other basic attributes, such as name and birthday, from pre-defined rules when they are not strongly constrained by visual evidence. 

\noindent\textbf{Step 3. Context-aware Attributes.} We prompt Gemini-2.5-Pro to generate attributes such as preferences and personal events, which later serve as major sources for question generation, conditioned on both the previously fixed attributes and the associated visual query. 

\noindent\textbf{Step 4. Group-wise Attributes.} We construct inter-concept relationships by randomly grouping four people, assigning shared attributes, and generating recurring events within the group, such as having dinner together every Tuesday. 

Through this structured pipeline, which combines model-assisted generation with rule-based construction, we aim to preserve attribute naturalness while improving the realism and diversity of the dataset.

The construction process for other concept types follows the same overall philosophy. In each case, we account for concept-specific characteristics in both attribute design and detailed generation procedures.

\begin{table*}[t]
\centering
\resizebox{0.99\textwidth}{!}{
\begin{tabular}{l ccccc>{\columncolor{gray!15}}c ccc>{\columncolor{gray!15}}c cccc>{\columncolor{gray!15}}c >{\columncolor{gray!15}}c}
\toprule
& \multicolumn{6}{c}{Vision-specific Reasoning}
& \multicolumn{4}{c}{Text-specific Reasoning}
& \multicolumn{5}{c}{Multimodal}
& \multicolumn{1}{c}{}
\\
\cmidrule(lr){2-7} \cmidrule(lr){8-11} \cmidrule(lr){12-16}
Method 
& \rotatebox{45}{predict} 
& \rotatebox{45}{spatial} 
& \rotatebox{45}{numeric} 
& \rotatebox{45}{commsense} 
& \rotatebox{45}{change} 
& \rotatebox{45}{V-Avg} 
& \rotatebox{45}{predict}
& \rotatebox{45}{numeric} 
& \rotatebox{45}{multi-hop} 
& \rotatebox{45}{T-Avg} 
& \rotatebox{45}{predict} 
& \rotatebox{45}{sentiment} 
& \rotatebox{45}{numeric}
& \rotatebox{45}{history}
& \rotatebox{45}{M-Avg}
& \rotatebox{45}{Avg}\\
\midrule
Random Choice 
& 25.0 & 25.0 & 25.0 & 25.0 & 25.0 & 25.0
& 25.0 & 25.0 & 25.0 & 25.0
& 25.0 & 25.0 & 25.0 & 25.0 & 25.0
& 25.0 \\

\midrule

\textbf{\textit{Proprietary+Prompting}} \\

GPT-4o~\cite{achiam2023gpt}+P
& 91.6 & 64.3 & 63.5 & 71.6 & 65.0 & 70.6
& 97.5 & 70.1 & 40.3 & 62.5
& 88.1 & 79.4 & 50.0 & 49.4 & 73.7
& 69.4 \\

GPT-5-mini~\cite{singh2025openai}+P
& 90.3 & 57.4 & 83.0 & 66.7 & 91.9 & 79.9
& 95.8 & 90.6 & 48.4 & 70.7
& 77.8 & 69.0 & 58.0 & 92.4 & 74.4
& 75.9 \\

GPT-5~\cite{singh2025openai}+P
& 88.4 & 66.1 & 81.2 & 72.6 & 87.8 & 80.3
& 98.3 & 95.8 & 77.4 & 87.2
& 88.1 & 79.4 & 64.0 & 94.9 & 84.1
& 83.1 \\

Gemini-2.5 Pro~\cite{comanici2025gemini}+P
& 85.8 & 62.6 & 77.7 & 63.7 & 94.3 & 77.9
& 98.3 & 95.8 & 83.0 & 90.0
& 84.1 & 83.3 & 59.5 & 96.2 & 85.1
& 83.1 \\

\midrule

\textbf{\textit{Open-sourced+Prompting}} \\

LLaVA~\cite{liu2024improved}+P 
& 69.7 & 46.1 & 50.2 & 51.9 & 90.2 & 60.2
& 61.7 & 41.7 & 18.6 & 35.5
& 34.9 & 49.9 & 20.0 & 39.2 & 40.3
& 48.2 \\

Qwen2.5-VL-7B~\cite{Bai2025Qwen25VLTR}+P
& 73.5 & 47.8 & 46.9 & 48.0 & 64.2 & 55.4
& 90.8 & 52.1 & 17.7 & 45.3
& 40.5 & 66.7 & 36.0 & 50.6 & 49.4
& 51.0 \\

\midrule

\textbf{\textit{Token-learning based}} \\

Yo'LLaVA~\cite{nguyen2024yo}  
& 47.7 & 28.7 & 33.9 & 47.1 & 42.3 & 39.0
& 43.3 & 30.2 & 14.0 & 25.6
& 30.1 & 34.9 & 30.0 & 25.3 & 31.9
& 33.5 \\

MC-LLaVA~\cite{an2024mc} 
& 56.1 & 39.9 & 43.5 & 39.2 & 65.0 & 48.4
& 57.5 & 35.4 & 13.6 & 30.4
& 39.7 & 50.0 & 34.0 & 25.3 & 39.6
& 41.3 \\

\midrule

\textbf{\textit{Retrieval based}} \\

RAP~\cite{hao2025rap} 
& 50.3 & 22.6 & 36.9 & 43.1 & 47.9 & 40.1
& 53.3 & 28.1 & 4.52 & 23.1
& 38.9 & 47.6 & 22.0 & 10.1 & 32.3
& 33.4 \\

R2P~\cite{das2025training}    
& 64.5 & 39.9 & 43.9 & 50.9 & 83.7 & 54.8
& 75.8 & 40.6 & 16.7 & 38.2
& 56.4 & 53.9 & 54.0 & 34.2 & 52.6
& 49.8 \\

\midrule

\textbf{\textit{Memory-augmentation-based}} \\

TAME~\cite{hong2026tameing}
& 72.9 & 52.2 & 50.6 & 51.9 & 77.2 & 59.8
& 89.2 & 61.5 & 9.05 & 42.6
& 60.3 & 61.1 & 46.0 & 31.7 & 50.2
& 52.6 \\

PersonaVLM~\cite{nie2026personavlm}
& 65.2 & 47.8 & 43.9 & 52.9 & 78.1 & 55.5
& 90.8 & 62.5 & 14.9 & 46.2
& 62.7 & 57.9 & 50.0 & 34.2 & 49.1
& 51.3 \\

\bottomrule
\end{tabular}}
\caption{\textbf{Evaluation on proposed MPRBench.}}
\vspace{-4mm}
\label{tab:main_table}
\end{table*}

\subsubsection{Query Construction}
\label{sec:query_construction}
After collecting the raw data and constructing the textual references, we use this information to build test instances, each consisting of a visual query, a textual query (i.e., question and answer choices), and the corresponding answer, for evaluating personalized models. During preliminary trials, we found that there is no single efficient and accurate strategy for constructing queries across all sub-tasks and concept types. A major reason is the large variation in data conditions, such as whether useful metadata (e.g., captions or bounding boxes) is available and whether sufficiently complex visual queries are abundant for a given sub-task. We therefore develop three types of query construction methods and choose among them depending on the available data conditions.

\noindent\textbf{Type 1: Template-based.}
For sub-tasks in which both useful metadata (e.g., captions or bounding boxes) and complex visual queries are available, we use templated questions and answer choices directly from the metadata. For example, this strategy is used for visual-specific tasks that leverage detailed captions in ConConChi and for numerical tasks that use bounding box annotations in RPC.

\begin{figure}[t]
    \centering
    \includegraphics[width=\linewidth]{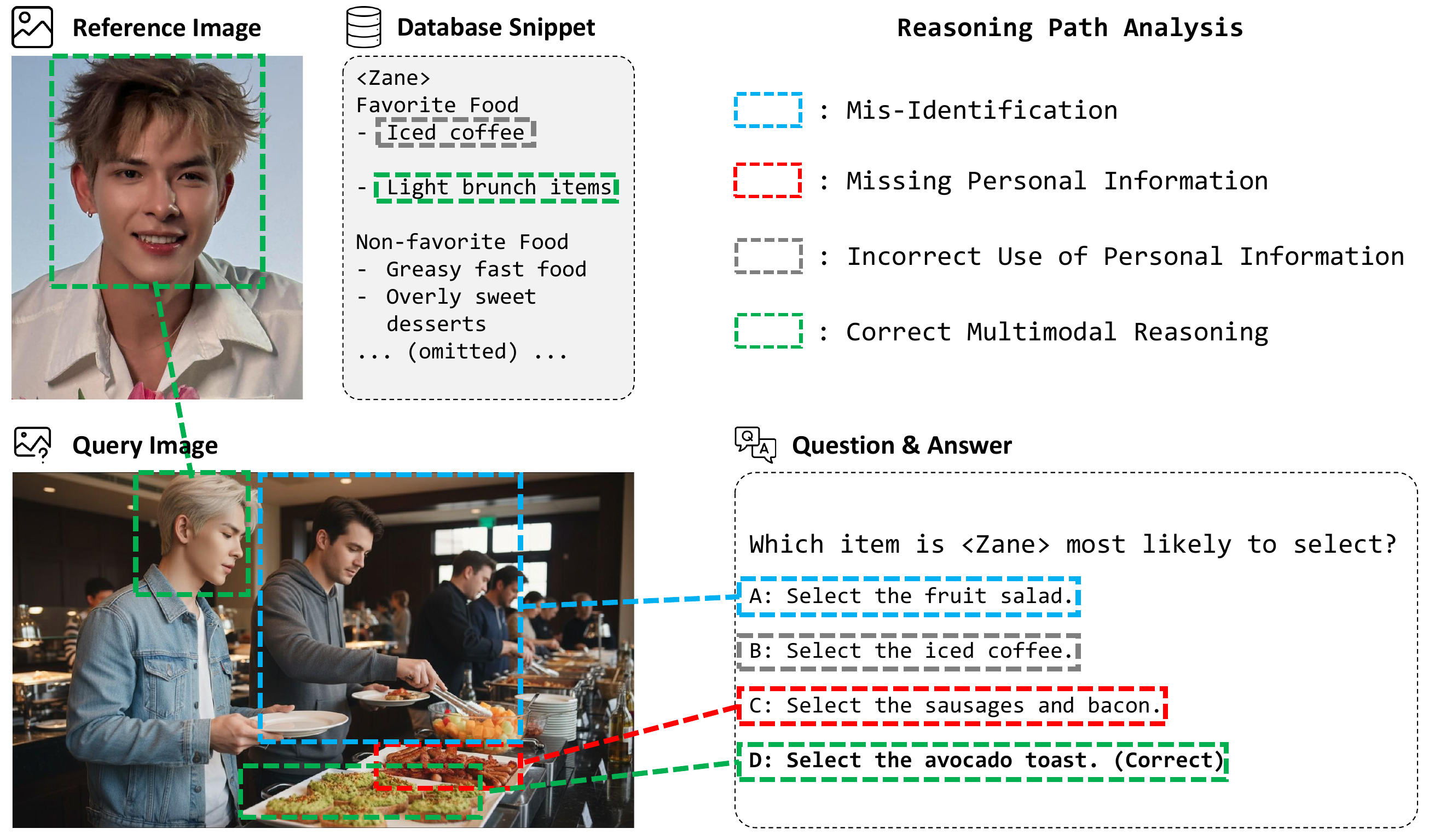}
    \caption{\textbf{Diagnostic Analysis of Multimodal Reasoning Paths}}
    \label{fig:failure_case}
\end{figure}

\noindent\textbf{Type 2: Human-labeled.}
When complex visual queries are available but supporting metadata is not, we manually construct the textual questions and answer choices. Annotators first identify the core reasoning logic of each problem by considering the nature of the sub-task, the visual query, and the textual reference. Based on this reasoning structure, they then craft appropriate questions and answer choices according to predefined guidelines. Most sub-tasks involving person concepts fall into this category.

\noindent\textbf{Type 3: Generative-model-based.}
In most cases, the above two methods allow us to construct high-quality queries. However, scaling up the dataset and increasing task difficulty remain challenging due to the scarcity of suitable complex visual queries. To address this limitation, we employ a state-of-the-art image generation model. Specifically, we first generate the core reasoning logic of the problem, followed by the textual question and answer choices, using Gemini-2.5-Pro. We then use Gemini 2.5 Flash Image~\cite{comanici2025gemini} to generate visual queries that match the generated textual questions. This strategy removes the dependence on pre-existing visual queries and allows us to better exploit the flexibility of generative models.

Although we employ three different query construction strategies, we maintain a common design philosophy of building an interpretable benchmark. To this end, regardless of the construction method, we design the incorrect answer choices to reflect the most likely failure cases of personalized models. This allows MPRBench to support comprehensive analysis of model behavior by aggregating statistics over different error types. In particular, the answer options are designed to capture representative failure types including: (1) \textbf{mis-identification}, where the model reasons as if another object or person were the target concept; (2) \textbf{missing personal information}, where the model falls back on generic commonsense instead of using the relevant personal information; and (3) \textbf{incorrect use of personal information}, where the model relies on personal information that is irrelevant to the given problem. As shown in Fig.~\ref{fig:failure_case}, each incorrect option is grounded in a distinct region of the query image or distinct entry of the personal database. This design allows the selected option to indicate which step of the reasoning process has failed.

\subsubsection{Human Verification}
To ensure the quality of MPRBench, we make substantial manual efforts to review the constructed data and filter out examples that are incorrect or ambiguous. This verification process includes checking the quality of intermediate outputs from raw data collection, textual reference construction, and query construction. \emph{Additional details on the human verification procedures used for quality control (\textit{e.g.,} annotator agreement across multiple aspects) are provided in the supplementary}.

\section{Personalized Reasoning Evaluation}

\subsection{Experimental Setup}
\label{sec:experimental_setup}
\noindent\textbf{Baseline models.}
We benchmark a wide range of personalization methods for MLLMs. These include: (1) \textbf{zero-shot, prompting-based personalization}, which injects personal information as additional context through prompting; (2) \textbf{personalized special-token learning methods}, which learn and use special tokens to represent the visual identity of personalized concepts, including Yo’LLaVA~\cite{nguyen2024yo} and MC-LLaVA~\cite{an2024mc}; (3) \textbf{retrieval-based models}, which employ similarity-based retrieval mechanisms for personalized concept recognition, such as RAP~\cite{hao2025rap} and R2P~\cite{das2025training}; and (4) \textbf{memory-augmentation-based models}, including TAME~\cite{hong2026tameing} and PersonaVLM~\cite{nie2026personavlm}. 
Note that memory-augmentation-based models are originally designed for long-term interactions across multiple turns. For evaluation on MPRBench, we repurpose them for single-turn evaluation by providing all concept-related information available in the benchmark as the model's single-turn memory input.

\noindent\textbf{Evaluation metrics.}
We report accuracy, defined as the proportion of samples for which a personalized model selects the correct answer, for each sub-task. To obtain representative metrics for each reasoning group (i.e., vision-specific, text-specific, and multimodal reasoning), we compute the accuracy across the sub-tasks within each group, weighted by the number of QA items in each sub-task. We denote these group-level metrics as V-Avg, T-Avg, and M-Avg, respectively. The final overall score, Avg., is defined as the weighted average of these three group-level scores.

\begin{figure}[t]
  \centering
  \includegraphics[width=0.90\linewidth]{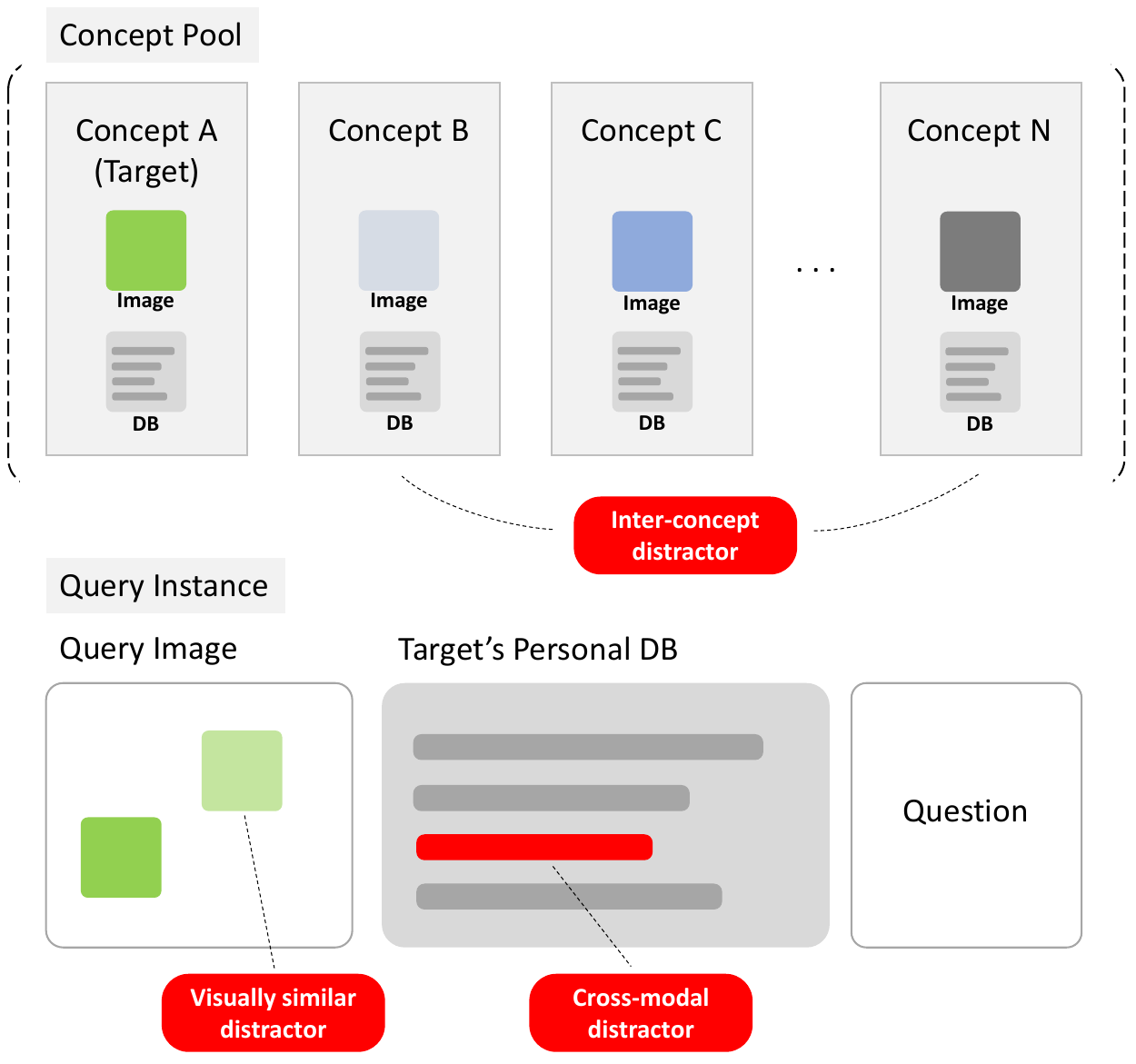}
  \caption{\textbf{Distractor Injection Points in the Input Structure.}}
  \label{fig:distractor}
  \vspace{-2mm}
\end{figure}

\subsection{Explored Various Distractors}
\label{sec:distractor}
To evaluate whether personalized MLLMs can selectively reason over target-relevant information, we explore various types of distractors introduced at different points in the model input, as illustrated in Fig.~\ref{fig:distractor}.

\noindent\textbf{Visually similar distractor.}
As described in Sec.~\ref{subsec:key}, we intentionally construct MPRBench with query images containing concepts visually similar to the target concept. If a model fails to localize the target accurately, it may instead ground its reasoning in a visually similar distractor and produce an answer based on the wrong entity.

\noindent\textbf{Cross-modal distractor.}
For vision-specific and text-specific tasks, where only one modality is required to answer correctly, we examine the effect of misleading information introduced through the other modality. Such information is superficially related to the query but irrelevant to the correct answer. We further investigate this distractor in the following section.

\noindent\textbf{Inter-concept distractor.}
Beyond the two distractors above, we expose models to reference images and personal databases of non-target concepts, either directly in the input for prompting-based models or in the concept pool for retrieval-based models. Most models, including proprietary ones, struggle to identify and reason over the target concept while ignoring these additional concepts, leading to performance degradation. We primarily analyze this distractor in the supplementary material.

\begin{figure}[t]
  \centering
  \includegraphics[width=0.98\linewidth]{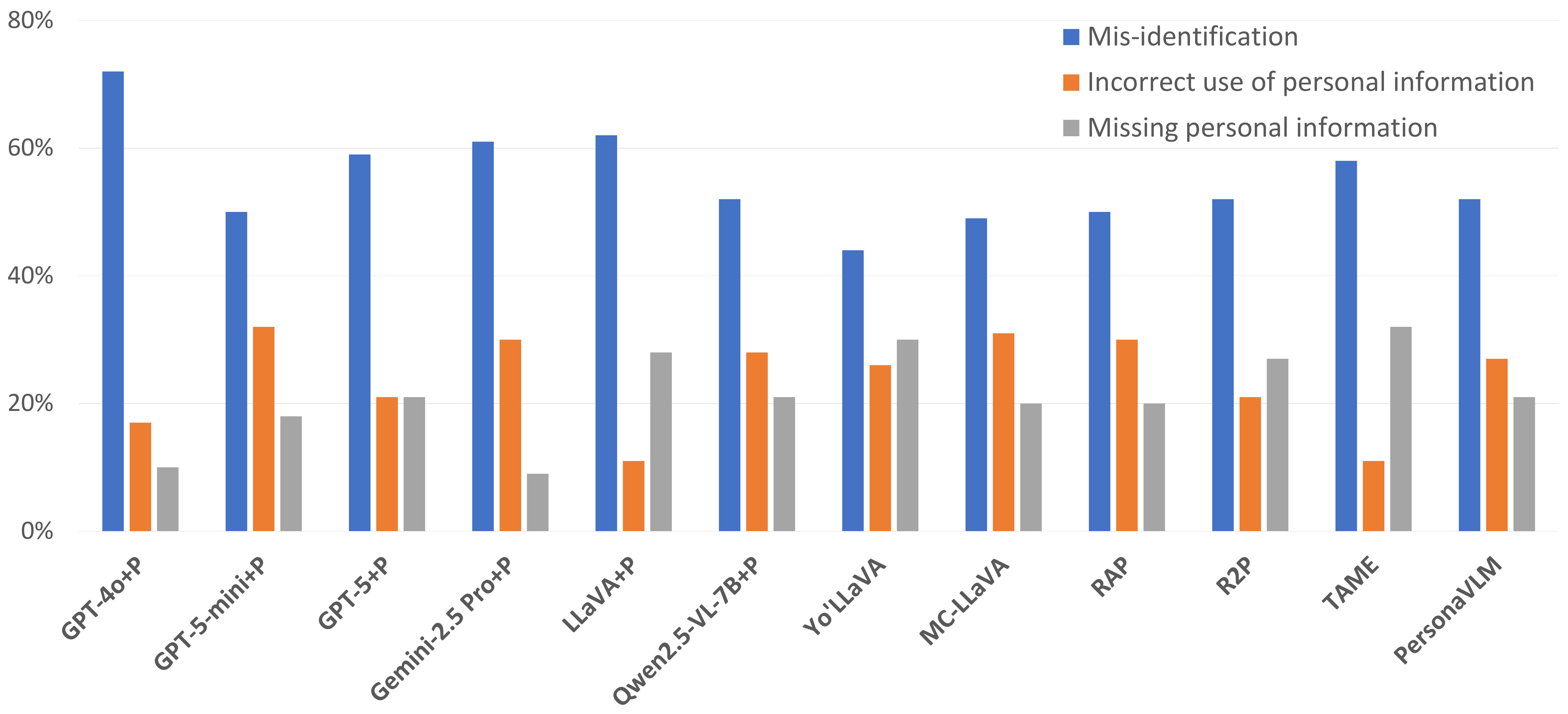}
  \vspace{-3mm}
  \caption{\textbf{Analysis on Failure Cases.}}
  \label{fig:error}
  \vspace{-4mm}
\end{figure}

\subsection{Evaluation Results and Analysis}
Based on MPRBench, we conduct extensive evaluations to assess the personalized reasoning capabilities of personalized MLLMs and to analyze the potential causes behind the observed performance trends.

\noindent\textbf{1) Can current personalized MLLMs perform personalized reasoning effectively?}
To answer this question, we benchmark baseline models on MPRBench (See \ref{tab:main_table}).
Overall, all baselines remain far from perfect, as reflected in the final Avg. score. This contrasts sharply with the near-perfect performance reported on previous personalized recognition benchmarks such as Yo’LLaVA, suggesting that \textit{current personalized MLLMs still struggle substantially with personalized reasoning tasks}.

More interestingly, methods specifically designed for personalization—both token-learning-based and retrieval-based approaches—often underperform a simple prompting baseline (i.e., LLaVA+Prompt). This observation contrasts with the strong performance claims made in prior works. In addition, RAP, which involves large-scale personalization tuning of the backbone MLLM, shows the weakest performance among the compared methods. We conjecture that tuning on elementary personalized tasks, such as recognition-oriented VQA, may improve personalized concept recognition to some extent, but such improvements do not generalize well to reasoning-intensive settings. One possible reason is that this type of tuning may not preserve, or may even weaken, the intrinsic reasoning capability of the original backbone model.

Memory-augmentation-based methods show only marginal gains over their Qwen2.5-VL-7B backbone with simple prompting. Although their tailored memory construction and retrieval pipelines improve retrieval over long-term histories, these gains do not directly translate to the reasoning required by MPRBench. In MPRBench, retrieved information must be further integrated with a cluttered visual context or commonsense knowledge through multi-step inference, going beyond simple retrieval.

\noindent\textbf{2) What are the common failure modes?}
Thanks to the interpretable option design of \texttt{MPRBench}, we can analyze model failures by summarizing the distribution of incorrectly selected answer choices. As shown in Fig.~\ref{fig:error}, the most common failure mode is mis-identification. This indicates that most models still struggle to accurately localize or identify the target concept among distractors during the reasoning process. We argue that this challenge has been somewhat overlooked in previous benchmarks, which typically evaluate personalized recognition in isolation and often rely on object-centric images. In realistic settings, however, personalized recognition itself becomes a major bottleneck for successful reasoning.

To further investigate the impact of personalized recognition on final performance, we conduct a controlled study comparing two settings: `reasoning only (location given)' versus `localization + reasoning'. For this experiment, we use the visual commonsense reasoning sub-task, since the VCR~\cite{zellers2019recognition} dataset provides precise bounding boxes for concepts. We manually select test samples for which the correct answer could change if mis-identification occurs. We then compare the performance of: (1) the LLaVA+Prompt baseline, where both localization and reasoning are required, and (2) a setting in which the target bounding box is provided as a visual prompt~\cite{cai2024vip}, so that only reasoning is required. As shown below, the large performance gap between these two settings indicates that personalized recognition remains a major challenge for reasoning-time performance.

\begin{center}\vspace{-.2em}
\tablestyle{4pt}{1.05}
\begin{tabular}{x{100}x{100}}
reasoning only & localization + reasoning \\
\shline
71.43\% & 51.43\% 
\end{tabular}
\end{center}

\begin{figure}[t]
  \centering
  \includegraphics[width=0.98\linewidth]{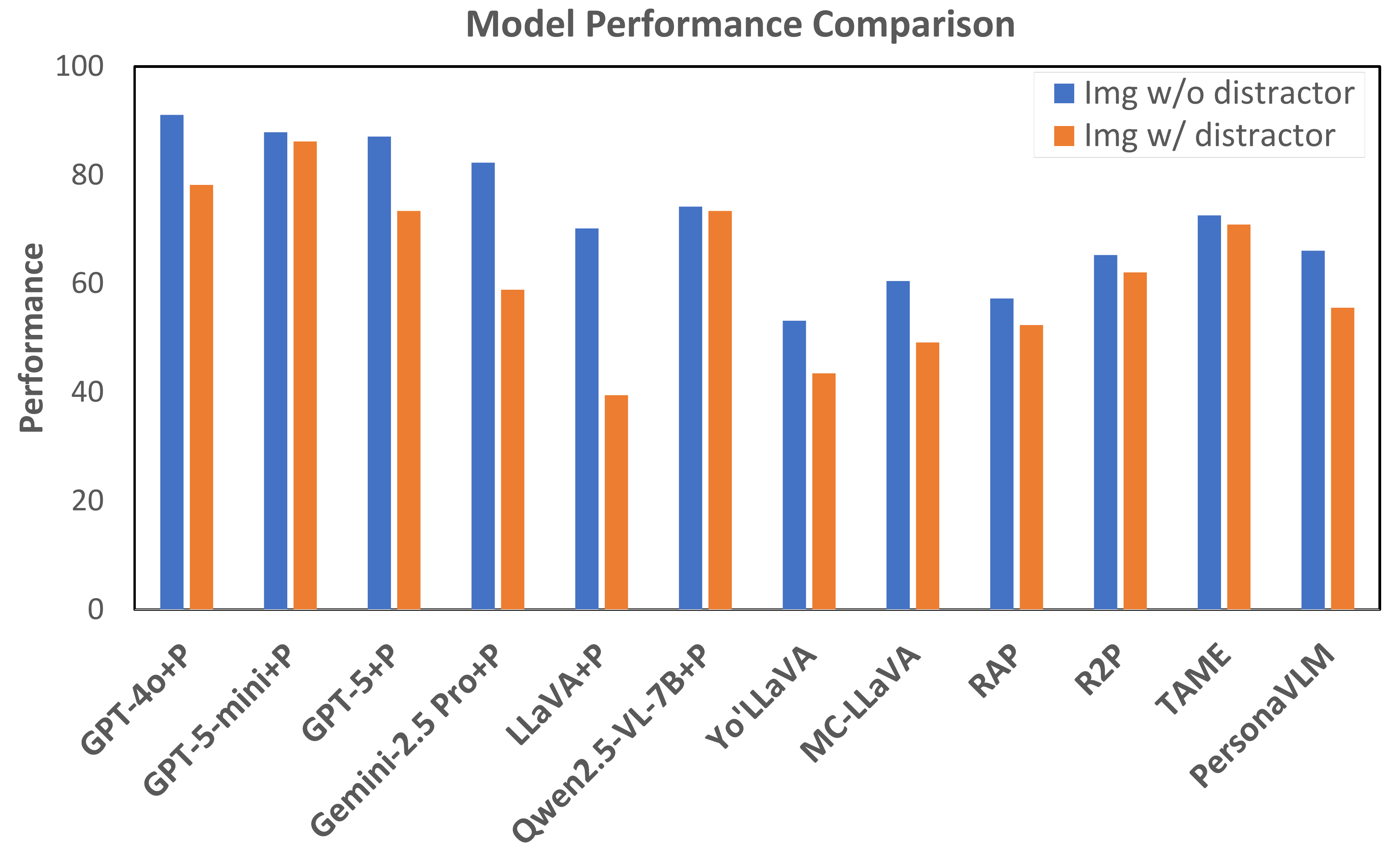}
  \vspace{-3mm}
  \caption{\textbf{Analysis on Cross-modal Distractor.}}
  \label{fig:distractor_analysis}
  \vspace{-4mm}
\end{figure}

\noindent\textbf{3) Are personalized MLLMs robust to cross-modal distractors?}
For the vision-specific and text-specific reasoning groups, it is important for models to reject irrelevant information from the other modality rather than relying on it during reasoning. This becomes particularly challenging when cross-modal personal information is irrelevant but still appears highly plausible or similar to the target concept. To investigate this issue, we conduct a controlled study in which we add hard negative textual personal information and measure its impact on model performance. We use the prediction task as the testbed for this analysis. Specifically, we introduce additional events using keywords taken from incorrect answer choices, while ensuring that the correct answer remains unchanged. As shown in Fig~\ref{fig:distractor_analysis}, most personalized MLLMs suffer noticeable performance drops when these distractors are added. This result suggests that \textit{current personalized MLLMs are not sufficiently robust to irrelevant textual distractors} and often fail to properly ignore misleading cross-modal personal information. 

\noindent\textbf{4) Which classes of methods excel at understanding fine-grained visual details of concepts?}
Understanding and leveraging fine-grained visual details of a personalized concept (e.g., hairstyle) is crucial in real-world scenarios. To examine which classes of methods are more effective in this setting, we focus on the visual appearance change reasoning task (“Change” in Table~\ref{tab:main_table}), which requires detailed comparison between the current visual context and the concept’s stored visual memory. In this task, token-learning-based methods generally perform worse than other three groups, all of which directly access the reference image itself. We suspect that compressing rich visual information into only a few learnable tokens inevitably causes information loss, leading to weaker performance on fine-grained visual understanding. 
Note that RAP is somewhat exception, which we suspect may stem from its tuning strategy weakening the model’s intrinsic reasoning capability.

\noindent\textbf{5) What unique challenges arise in personalized multi-hop reasoning?}
This task introduces the unique challenge of repeatedly leveraging personal information across multiple reasoning steps. Among all sub-tasks, personalized multi-hop reasoning is the most challenging for open-source personalized MLLMs, with most models performing at or even below random chance. Moreover, the performance gap between open-source models and frontier proprietary models is the largest across all sub-tasks, highlighting a critical limitation and the urgent need for more effective methods to support reasoning across multiple retrieval and inference steps.

\section{Conclusion}
In this paper, we introduce MPRBench, a novel benchmark for multimodal personalized reasoning. MPRBench consists of 12 sub-tasks spanning vision-specific, text-specific, and multimodal reasoning, providing broad coverage of personalized reasoning scenarios. In addition, we adopt an interpretable answer design that enables analysis of representative failure modes.
Using MPRBench, we find that current personalized MLLMs still face substantial challenges in personalized reasoning, particularly in accurate personalized recognition and robustness to cross-modal distractors during the reasoning process. We hope that our benchmark and analyses will provide useful insights for the development of future personalized MLLMs.

\section*{Acknowledgement}
This research was supported by the National Research Foundation of Korea(NRF) grant funded by the Korea government(MSIT)(RS-2026-25499205, 80\%). Also, the authors acknowledge the Urban Big Data and AI Institute of the University of Seoul supercomputing resources (http://ubai.uos.ac.kr) made available for conducting the research reported in this paper (20\%).

{
    \small
    \bibliographystyle{ieeenat_fullname}
    \bibliography{main}
}

\clearpage
\appendix
\section*{Overview}
\label{sec:supp_overview}

This supplementary material reports the dataset statistics, the full construction procedure and the additional experimental results of MPRBench. \\

\begin{itemize}
  \item \textbf{Sec.~\ref{sec:mprbench}} provides more details on MPRBench itself.
  \begin{itemize}
  \item \textbf{Sec.~\ref{sec:benchmark_statistics}} reports detailed benchmark statistics.
  \item \textbf{Sec.~\ref{sec:details_textual}} provides the full textual reference construction process and \textbf{examples of personal database}.
  \item \textbf{Sec.~\ref{sec:details_query}} describes the query construction pipeline.
  \item \textbf{Sec.~\ref{sec:details_verification}} reports the human verification procedure and agreement statistics, with the annotation interface used.
  \item \textbf{Sec.~\ref{sec:details_task}} provides the definition and \textbf{representative examples} of all 12 sub-tasks. \\
  \end{itemize}
  
  \item \textbf{Sec.~\ref{sec:details_experiments}} provides additional experimental details and analyses.
  \begin{itemize}
      \item \textbf{Sec.~\ref{sec:details_setup}} describes how each baseline method was adapted to MPRBench.
      \item \textbf{Sec.~\ref{sec:analysis_real_syn}} compares model performance on naturally collected versus synthetically generated visual queries.
      \item \textbf{Sec.~\ref{sec:details_distractor}} provides analysis of various distractor types.
      \item \textbf{Sec.~\ref{sec:experiments_qwen}} provides further experiments with Qwen3-VL family models.
  \end{itemize}
\end{itemize}

\section*{Limitations}
Given the complexity and diversity of multimodal personalized reasoning, MPRBench may not cover all possible scenarios. Nevertheless, we make a substantial effort to include a wide range of sub-tasks and to introduce novel tasks in each modality-specific group. Exploring even more diverse and complex scenarios for multimodal personalized reasoning remains an important direction for future work.

In addition, our primary goal in this work is to benchmark and analyze the current multimodal reasoning capabilities of personalized MLLMs. As a result, we do not extensively investigate methodological approaches for improving model performance. However, we believe that the findings and analyses provided in this work can serve as a useful foundation for future advances in personalized reasoning.

\section*{Ethics Statement}
Because our benchmark includes person concepts and images of people, we strictly follow the licenses and terms of use associated with all source images. Most images are drawn from open datasets, and the remaining images are collected under CC BY 4.0 licenses. In particular, we carefully verify whether each person image permits derivative use or editing. If editing is not allowed, we do not use that image as a basis for generated or modified content (e.g., in Type 3 query construction).

For textual references, namely the personal information associated with images, we intentionally generate synthetic attributes rather than relying on factual personal information such as real names or actual biographies. In this way, we aim to mitigate privacy concerns.

In preparing this manuscript, we used ChatGPT~\cite{singh2025openai} to improve readability by checking grammar and polishing language. All core content, ideas, and claims were originally written by the authors, who take full responsibility for the contents of the paper.

\section{More Details on MPRBench}
\label{sec:mprbench}

\subsection{Benchmark Statistics}
\label{sec:benchmark_statistics}
\begin{figure*}[t]
    \centering
    \includegraphics[width=\linewidth]{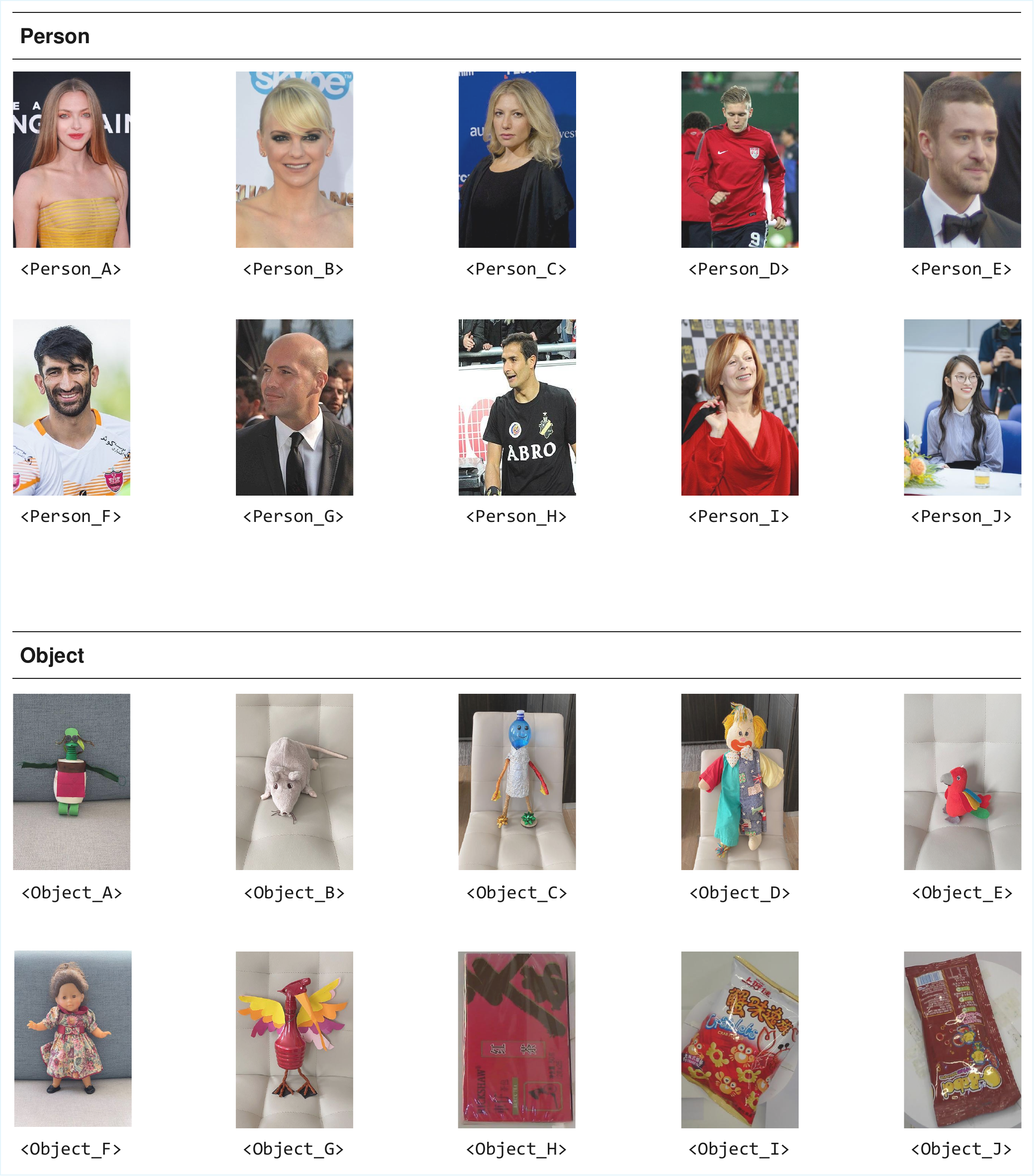}
    \caption{\textbf{Overview of MPRBench Personalized Concepts}}
    \label{fig:concept_overview}
\end{figure*}

In this section, we provide a detailed numerical analysis of MPRBench to highlight its scale and the diversity of its evaluation scenarios. \\
\textbf{Diversity across Personalized Concepts}: As summarized in Table.~\ref{tab:concept_stats}, our benchmark incorporates a wide array of 119 unique personalized concepts, comprising 69 distinct individuals and 50 varied objects. For person concepts, we draw from four existing personalization datasets, from SoccerAgent, and from CC BY 4.0 licensed web images; for object concepts, we repurpose ConconChi~\cite{rosasco2024concon} and RPC~\cite{wei2019rpc}, which were originally built for object retrieval and retail product checkout. Fig.~\ref{fig:concept_overview} provides an overview of selected personalized concepts used in our benchmark. This extensive collection of concepts ensures that the models are evaluated on their ability to generalize across diverse visual identities. \\
\textbf{Broad Coverage of Reasoning Tasks}: Furthermore, Table.~\ref{tab:task_stats} details the distribution of 1,667 question-answer pairs across 12 sub-tasks. By categorizing these to visual-specific (766 QAs), text-specific (437 QAs), and Multimodal (464 QAs), MPRBench covers a broad spectrum of reasoning requirements. This comprehensive task design ensures a rigorous assessment of a model's multifacted capabilities.

\begin{table}[t]
\centering
\small
\begin{tabular}{llr}
\toprule
\textbf{Category} & \textbf{Source} & \textbf{\# Concepts} \\
\midrule
\multirow{7}{*}{Person}
 & Yo'LLaVA~\cite{nguyen2024yo}            & 7 \\
 & MC-LLaVA~\cite{an2024mc}                & 9 \\
 & MMPB~\cite{kim2026mmpb}                 & 6 \\
 & CustomConcept101~\cite{kumari2023multi} & 3 \\
 & SoccerAgent~\cite{rao2025multi}         & 22 \\
 & Web                                     & 22 \\
\cmidrule{2-3}
 & \textit{Subtotal}                       & \textit{69} \\
\midrule
\multirow{3}{*}{Object}
 & ConConChi~\cite{rosasco2024concon}     & 20 \\
 & RPC~\cite{wei2019rpc}                   & 30 \\
\cmidrule{2-3}
 & \textit{Subtotal}                       & \textit{50} \\
\midrule
\multicolumn{2}{l}{\textbf{Total}}         & \textbf{119} \\
\bottomrule
\end{tabular}
\caption{Distribution of personalized concepts in \texttt{MPRBench} across concept categories and source datasets.}
\label{tab:concept_stats}
\end{table}

\begin{table}[ht]
\centering
\small
\captionsetup{justification=centering}
\renewcommand{\arraystretch}{1.2}
\begin{tabular}{l l r} 
\toprule
\textbf{Task Category} & \textbf{Sub-task} & \textbf{\# Questions} \\ \midrule

\textbf{Visual-Specific} & Prediction & 155 \\
& Spatial & 115 \\
& Numerical & 271 \\
& Commonsense & 102 \\
& Visual Appearance Change & 123 \\ \cmidrule(lr){2-3}
& \textit{Sub-total (Visual)} & \textbf{766} \\ \midrule

\textbf{Text-Specific} & Prediction & 120 \\
& Numerical & 96 \\
& Multi-Hop & 221 \\ \cmidrule(lr){2-3}
& \textit{Sub-total (Text)} & \textbf{437} \\ \midrule

\textbf{Multimodal} & Prediction & 209 \\
& Sentiment & 126 \\
& Numerical & 50 \\
& history & 79 \\ \cmidrule(lr){2-3}
& \textit{Sub-total (Multimodal)} & \textbf{464} \\ \midrule[\heavyrulewidth] 

\multicolumn{2}{l}{\textbf{Total}} & \textbf{1,667} \\ \bottomrule
\end{tabular}
\caption{Detailed statistics of MPRBench across different task categories.}
\label{tab:task_stats}
\end{table}

\begin{table*}[ht]
\centering
\small
\renewcommand{\arraystretch}{1.4}
\begin{tabularx}{\textwidth}{l l X}
\toprule
\textbf{Category} & \textbf{Attribute} & \textbf{Definitions} \\ \midrule

\multirow{3.5}{*}{Visually-grounded} 
    & gender & Gender of Person Concept which is strongly correlated with the concept's appearance in reference images. \\ \cline{2-3}
    & height & Height of Person Concept which inferred from visual context and diversity-oriented rules. \\ \midrule

\multirow{4.5}{*}{Diversity-oriented} 
    & name & Name of Person Concept by using unique name list to avoid duplication. \\ \cline{2-3} 
    & birthday & Birthday of Person Concept randomly generated based on estimated age to ensure temporal diversity. \\ \cline{2-3}
    & family & Family Information about Person Concept which includes Family Relation with its birthday each. Randomly generated based on Person Concept's birthday and Replationship. \\ \midrule

\multirow{10.5}{*}{Context-aware} 
    & personality & Personality of Person Concept generated based on visual queries and fixed attributes. \\ \cline{2-3}
    & favorites/non-favorites & Favorite/non-favorite of Person Concept which includes favorite/non-favorite colors, foods, and fashions. \\ \cline{2-3}
    & hobby & Hobby of Person Concept which includes Personal interests and activities generated based on visual queries. \\ \cline{2-3}
    & habit & Habit of Person Concept which includes not only visual information but also textual information so that it facilitates question generation and effectively reflects complex factors such as real-world routines. \\ \cline{2-3}
    & event & Event of Person Concept includes personal events prevent in the visual query with the plausible behind story which is generated. Similar to habit, it facilitates question generation and effectively reflects complex real-world factors such as date, emotion, and action. \\ \cline{2-3}
    & nickname & Nickname of Person Concept which can be used instead of their name when referring to them implicitly. \\ \midrule

\multirow{8.5}{*}{Group-wise}
    & birthplace & Birthplace of Person Concept which is shared with group members. Since it is consistent across group members and is unrelated to bias based on visual information - randomly assigned by city list, it can act as a hard negative element.\\ \cline{2-3}
    & occupation & Occupation of Person Concept which is randomly assigned by list. \\ \cline{2-3}
    & residence & Residence of Person Concept which is randomly assigned by list. \\ \cline{2-3}
    & group\_event & Event between Person Concept and group members which is partitioned among members and the events can be connected. For example, They meet every Monday - Every Monday they have dinner together. This is neccessary when solving multi-hop question. The format of events are same but each elements(e.g., "Monday", "dinner") are randomly assigned by list. \\ 
\bottomrule
\end{tabularx}
\caption{Full List and Definitions of Attributes in the Textual Reference for Person Concepts.}
\label{tab:attribute_words_person}
\end{table*}

\subsection{Details on Textual Reference Construction}
\label{sec:details_textual}
In this section, we provide the full list of attributes used in the textual references, together with their definitions. Tables ~\ref{tab:attribute_words_person} and ~\ref{tab:attribute_words_objects} summarize this information for person and object concepts, respectively. While the construction process for person concepts is detailed in Section~\ref{sec:person_construction}, this section provides the process for object concepts. In contrast to person concepts which incorporate biological attributes (e.g., age, gender), the construction of object concepts focuses on ownership mapping and history, as detailed below.

\noindent\textbf{Step 1. Identification Attributes.} Since object concepts lack biological attributes, they are instead initialized with unique identifiers to ensure distinctiveness. Each object is assigned a specific name from a predefined unique name list (e.g., for Conconchi~\cite{rosasco2024concon} or RPC~\cite{wei2019rpc}) and an \texttt{<owner>} placeholder. This placeholder serves as a functional link, allowing the object to be dynamically associated with a specific person concept in later stages. 

\noindent\textbf{Step 2. Diversity-oriented Attributes.} To promote diversity, we randomly assign temporal and ownership attributes, such as acquisition date, manufacturing date, and the actual owner. These attributes are generated based on pre-defined rules to maintain logical consistency (e.g., manufacturing date preceding acquisition date). 

\noindent\textbf{Step 3. Context-aware Attributes.} We prompt Gemini-2.5-Pro~\cite{comanici2025gemini} to extract or generate event-based attributes by analyzing the query images. For Conconchi concepts, the model generates a logical chain of information grounded in visual evidence. In contrast, for RPC, where query images serve for numerical tasks, these attributes are generated to maintain context-aware consistency. These attributes include external textual facts, such as specific dates and personal events, which are non-inferable from the image alone.
In designing these attributes, we aim to account for the characteristics specific to each concept type. The full prompts used at model-assisted steps are provided in Figs.~\ref{fig:construction_prompt_person_step1} to~\ref{fig:construction_prompt_person_step3} for person concepts and Figs.~\ref{fig:construction_prompt_step3_object_start} to~\ref{fig:construction_prompt_step3_object_end} for object concept, respectively. We provide examples of the personal databases in Figs.~\ref{fig:database_person} and~\ref{fig:database_object}, demonstrating the structured textual information that models must utilize alongside visual inputs to solve the reasoning tasks in MPRBench.

\begin{table*}[ht]
\centering
\small
\renewcommand{\arraystretch}{1.5}
\begin{tabularx}{\textwidth}{l l X}
\toprule
\textbf{Category} & \textbf{Attribute} & \textbf{Definitions} \\ \midrule

\multirow{1}{*}{Identification} 
    & name & Name of Object Concept assigned by a unique name list to avoid duplication. \\ \midrule

\multirow{3.5}{*}{Diversity-oriented} 
    & acquisition\_date & Acquisition date of Object Concept. \\ \cline{2-3}
    & manufacturing\_date & Manufacturing date of Object Concept which is always before acquisition date. \\ \cline{2-3}
    & owner & Name of Object Concept's Owner which assigned Person Concept Name for further extensibility . \\ \midrule

\multirow{7.5}{*}{Context-aware} 
    & object\_type & Type of Object Concept. \\ \cline{2-3}
    & acquisition\_process & Acquisition process of Object Concept. \\ \cline{2-3}
    & history & History of Object Concept which has similar composition with Person Concept's event attribute. Contrastingly history can includes interaction history with its owner(does not serve as "Person Concept" but "owner"). To generate multimodal problems of high difficulty and ensure that the length of the text is similar to that of the Person Concept's Textual Reference, a distractor history---similar to but distinct from the history extracted from the visual query---is generated, and the history is created as a pair. \\ \cline{2-3}
    & storage & Storage place of Object Concept. \\ 
\bottomrule
\end{tabularx}
\caption{Full List and Definitions of Attributes in the Textual Reference for Object Concepts.}
\label{tab:attribute_words_objects}
\end{table*}

\subsection{Details on Query Construction}
\label{sec:details_query}
In this section, we provide the detailed mapping between sub-tasks and query construction types in the table~\ref{tab:mapping_query} described in Sec.~\ref{sec:query_construction}. This mapping is determined by the source of the underlying concept rather than by the sub-task alone. Object concepts from ConConChi and RPC carry captions and bounding box annotations, therefore admit template-based construction (Type 1). Person concepts lack such metadata, so their queries are manually constructed instead (Type 2). Type 3 is used across both concept types when suitable complex visual queries are scarce. Figs.~\ref{fig:prompt_step1_person} to ~\ref{fig:prompt_step4_person} provide the prompts used for Type 3 construction, illustrated with the predictive reasoning sub-task of person concepts. We first design a scene blueprint that places a favorite, a non-favorite, and a neutral item in the same scene together with a distractor person, then convert this blueprint into a question and four answer choices, and finally generate the corresponding visual query while preserving the identity of the target concept. The remaining Type 3 sub-tasks follow the same four-stage procedure, and differ only in the blueprint rules of the first stage.

\subsection{Details on Human Verification}
\label{sec:details_verification}
\begin{figure*}[hb]
    \centering
    \includegraphics[width=\linewidth]{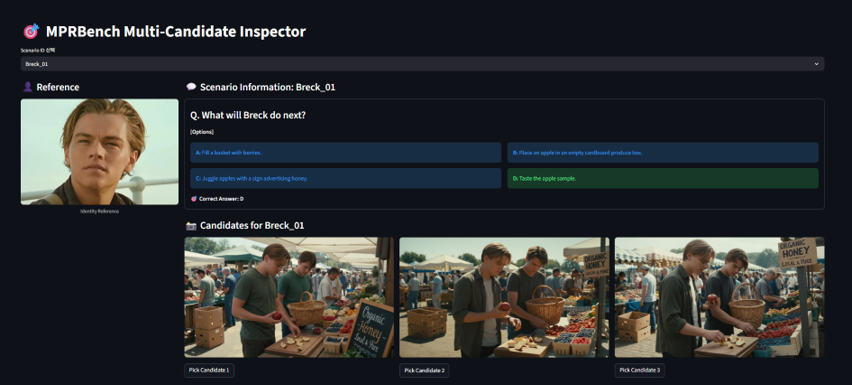}
    \caption{\textbf{Human Verification Case.}}
    \label{fig:human_verify}
\end{figure*}

In this section, we describe the human verification process conducted to ensure the high quality and complexity of MPRBench. This verification covers both the generated textual references and the collected or generated visual queries. Fig.~\ref{fig:human_verify} illustrates the interfaces used for human verification.

\noindent\textbf{Validation of generated textual references.} Since parts of the personal database are model-generated, we validate them along criteria appropriate to each attribute type, and report the results in Table.~\ref{tab:text_validation}. For visually grounded attributes, we treat the reference image as ground truth. Gender predictions were consistent with both annotators in all inspected cases, and age predictions fell within a 10-year range of the annotator's estimates in 96.0\% of cases. For subjective attributes such as preferences, there is often no single correct answer, and many generated values are equally plausible. The relevant criterion is therefore internal consistency rather than uniqueness, and annotators manually cross-referenced the generated color, food, and fashion preferences to eliminate contradictions, such as an item being listed as both a favorite and a non-favorite for the same concept; the two annotators reached the same decision in 99.3\% of cases. For event attributes, we require that each event neither contradicts other attributes of the same concept nor conflicts with the associated visual query, so that valid question-answer pairs can be constructed on top of it. Candidate events violating either requirement were discarded, with an agreement rate of 94.0\%. Finally, for group-wise attributes, we inspected relational integrity to confirm that shared attributes such as residences and events were consistently mapped across all associated members.

\begin{table}[t]
\centering
\small
\begin{tabular}{llcc}
\toprule
\textbf{Attribute} & \textbf{Validation criterion} & $N$ & \textbf{Agr.} \\
\midrule
Gender     & match w/ annotators       & 50  & 100.0 \\
Age        & within $\pm$10-year range & 50  & 96.0 \\
Preference & free of contradiction     & 150 & 99.3 \\
Event      & consistent \& grounded    & 150 & 94.0 \\
\bottomrule
\end{tabular}
\caption{Validation of the generated textual references. $N$ denotes the number of inspected items, and Agr.\ denotes the agreement rate (\%) between two annotators.}
\label{tab:text_validation}
\end{table}

\begin{table}[t]
\centering
\small
\begin{tabular}{lccc}
\toprule
\textbf{Stage} & \textbf{Inspected} & \textbf{Retained} & \textbf{Rate} \\
\midrule
VCR query selection    & 4{,}000 & 120 & 3.0\% \\
Type-3 image filtering & 3{,}117 & 579 & 18.6\% \\
\bottomrule
\end{tabular}
\caption{Retention rates of the visual query filtering stages in \texttt{MPRBench}.}
\label{tab:visual_filtering}
\end{table}

\noindent\textbf{Filtering of visual queries.} We further apply a systematic filtering process to visual queries, summarized in Table.~\ref{tab:visual_filtering}. As discussed in Sec.~\ref{sec:raw_data_collection}, we prioritize scene complexity in order to evaluate concept identification under challenging conditions. For human-centric tasks, we manually reviewed 4,000 candidate images from VCR and retained only 120 that satisfied our criteria of high scene density, presence of distractors, and cluttered backgrounds. For Type-3 queries, we found that current image generation models rarely satisfy all of our requirements at once, namely preserving the identity of the personalized concept, reflecting the intended number of target concepts and distractors, and maintaining the naturalness of the overall scene. We therefore generated three to five candidates per scenario and retained only the best one, resulting in 579 images out of 3,117 candidates.

Through this comprehensive manual effort, we ensured that MPRBench remains a reliable and logically sound instrument for evaluating multimodal personalized reasoning.

\subsection{Comprehensive Task Definitions}
\label{sec:details_task}
\begin{figure*}[t]
    \centering
    \includegraphics[width=0.88\linewidth]{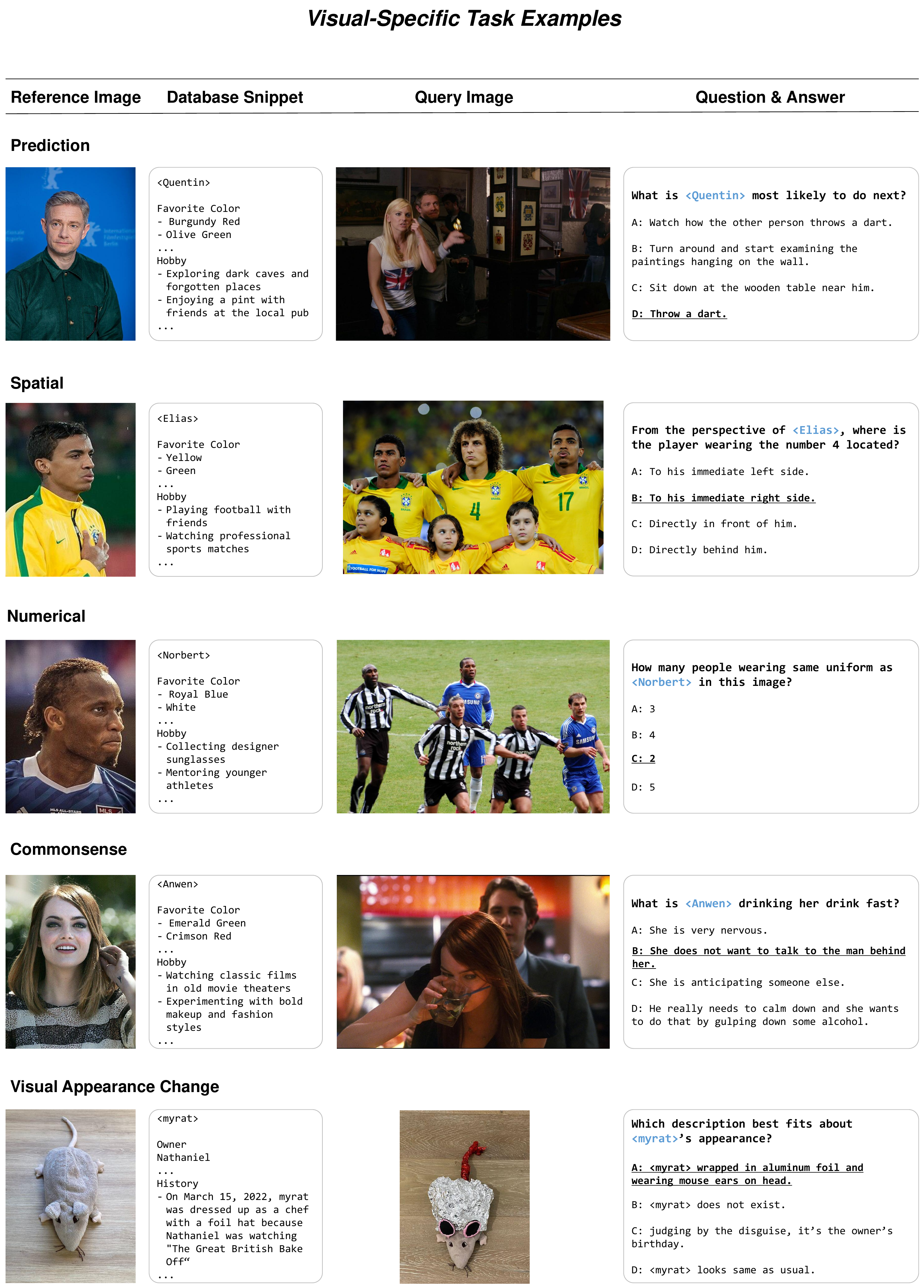}
    \caption{\textbf{Data Examples for Visual-Specific Personalized Reasoning}}
    \label{fig:visual-specific}
\end{figure*}

\begin{figure*}[t]
    \centering
    \includegraphics[width=0.88\linewidth]{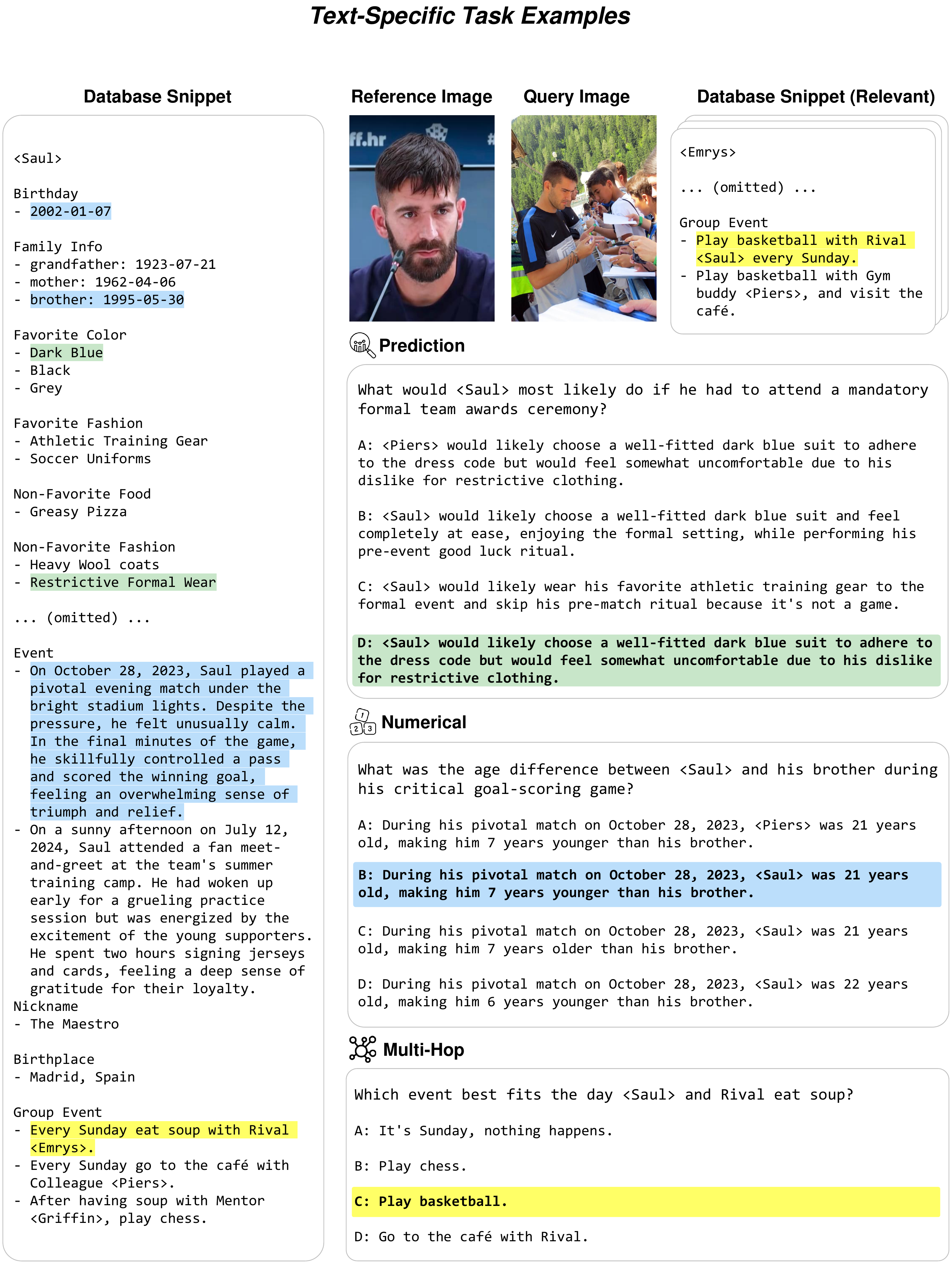}
    \caption{\textbf{Data Examples for Text-Specific Personalized Reasoning}}
    \label{fig:text-specific}
\end{figure*}

\begin{figure*}[t]
    \centering
    \includegraphics[width=0.88\linewidth]{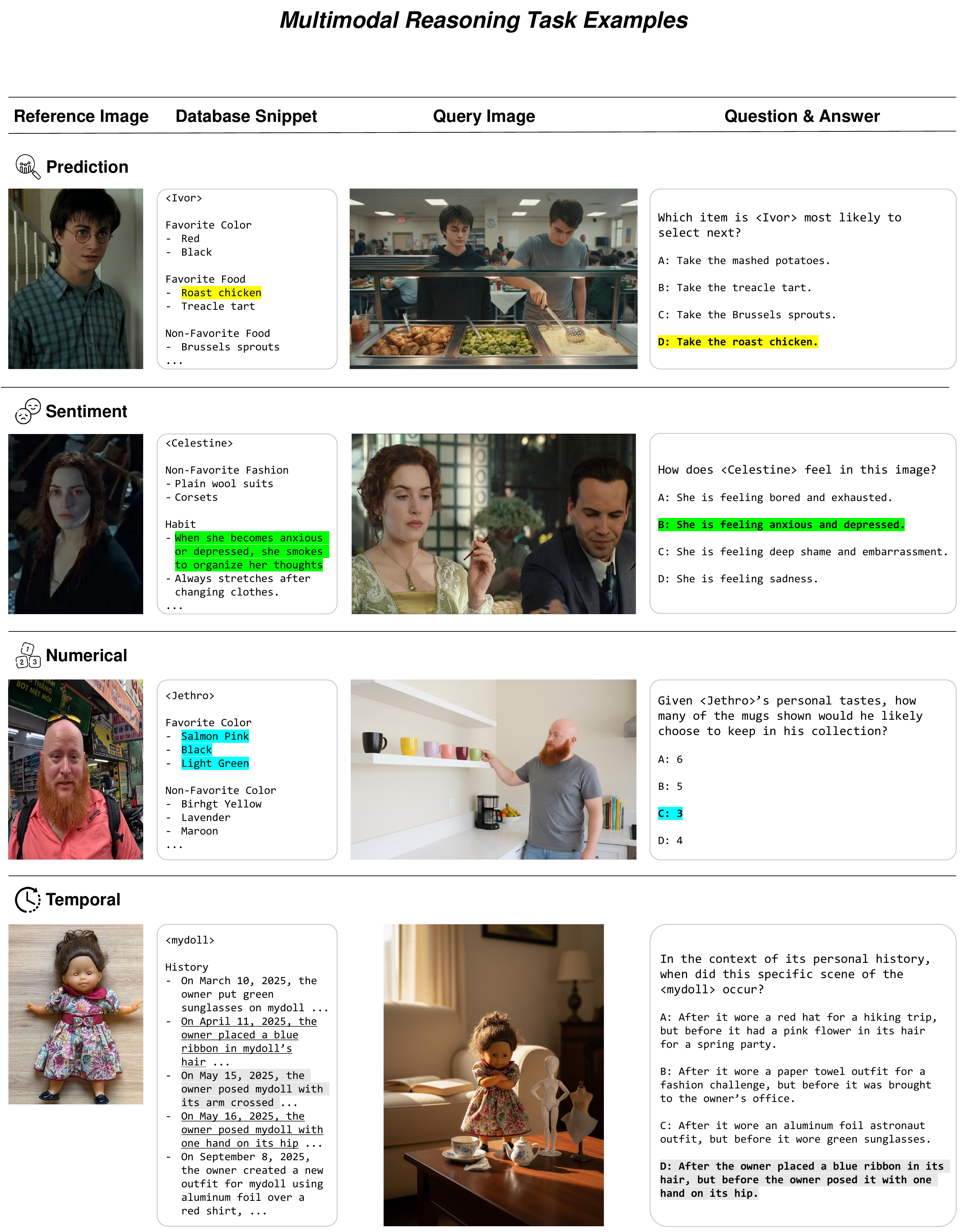}
    \caption{\textbf{Data Examples for Multimodal Personalized Reasoning}}
    \label{fig:mm-examples}
\end{figure*}

Sec.~\ref{sec:task_definition} introduces the 12 sub-tasks and the reasoning demands they impose. In this section, we specify the concrete form each sub-task takes in MPRBench, i.e., what the question asks and what evidence a correct answer requires. Figs.~\ref{fig:visual-specific} to~\ref{fig:mm-examples} illustrates the examples of each sub-task.

\noindent\textbf{Visual-Specific Reasoning.} The correct answer is determined entirely by the query image, and the personal database contributes only the identity of the target concept.
\begin{itemize}
    \item \textbf{Prediction Reasoning:} The scene presents several courses of action available to the concept, and the question asks which one follows from its current pose and surroundings.
    \item \textbf{Spatial Reasoning:} Positional relations are asked from the concept's own perspective rather than from the viewer's, so that the concept must be localized before the relation can be resolved.
    \item \textbf{Numerical Reasoning:} The question asks how many entities in the scene share a given visual attribute with the concept, such as wearing the same uniform.
    \item \textbf{Visual Commonsense Reasoning:} The question asks why the concept is in behaving or appearing a certain way, and the options offer distinct commonsense explanations, only one of which is consistent with the visual context.
    \item \textbf{Visual Appearance Change Reasoning:} The question asks how the concept's current appearance differs from the reference image(s).
\end{itemize}

\noindent\textbf{Text-Specific Reasoning.} Reference and query images are provided, but the correct answer is fully determined by the personal database.
\begin{itemize}
    \item \textbf{Prediction Reasoning:} The question poses a hypothetical condition that is absent from the database, and the answer must be derived from recorded preferences, habits, or past events.
    \item \textbf{Numerical Reasoning:} The question requires arithmetic over dated attributes, such as computing an age difference at the time of a recorded event.
    \item \textbf{Personalized Multi-Hop Reasoning:} The answer is not stated in any single entry and must be assembled by chaining group-wise attributes across two or more concepts.
\end{itemize}

\noindent\textbf{Multimodal Reasoning.} The correct answer requires both sources; neither the image nor the database alone is sufficient.
\begin{itemize}
    \item \textbf{Prediction Reasoning:} The question asks what the concept is likely to do next, and the answer must be inferred by combining multiple elements present in the query image with relevant attributes recorded in the database.
    \item \textbf{Sentiment Reasoning:} The question asks what emotion the concept is feeling, or why, and the answer must be inferred by jointly considering the query image and relevant attributes recorded in the database.
    \item \textbf{Numerical Reasoning:} The question asks how many objects in the scene satisfy a criterion stated in the database, so counting depends on a textual attribute.
    \item \textbf{Multimodal history Reasoning:} The question asks when the query image was taken, and the answer is the position in the concept's recorded history whose described visual state matches the scene.
\end{itemize}

\section{More Details on Experiments}
\label{sec:details_experiments}
\subsection{Experimental Setup}
\label{sec:details_setup}
In this section, we describe how we adapted various MLLM baselines to the MPRBench framework. \\
\textbf{Prompting-based Personalization (Zero-shot)}: For general MLLMs like GPT-4o~\cite{achiam2023gpt}, Qwen2.5-VL~\cite{Bai2025Qwen25VLTR} and LLaVA-v1.5~\cite{liu2023visual}, we provided personal information as structured context. Specifically, we converted the personal database into a JSON-formatted string and inserted it into the prompt. This ensures that the model can parse the attributes (e.g., identity, preferences, history) in a standardized manner. \\
\textbf{Special-token Learning Methods (Yo'LLaVA~\cite{nguyen2024yo}, Mc-LLaVA~\cite{an2024mc})}: These models utilize learnable tokens to represent specific identities. Both models were implemented using LLaVA-v1.5-13B as the base architecture. For evaluation on MPRBench, we followed the original training protocols described in their respective papers to learn the special tokens for each concept. During the evaluation phase, we provided the corresponding personal database via structured prompting as a dynamic context. \\
\textbf{Retrieval-based Models (RAP~\cite{hao2025rap}, R2P~\cite{das2025training})}: We utilized the official source code and pre-trained checkpoints provided by the original authors. RAP is built on the LLaVA-v1.5-13B backbone, while R2P is based on the MiniCPM-o-2.6 architecture. To adapt them to our benchmark, we integrated the personalized history for each concept from our database into their respective concept pools, so that the model retrieves the concept by visual similarity and conditions on the profile stored with it. \\
\textbf{Memory-augmentation-based models (TAME~\cite{hong2026tameing},  PersonaVLM~\cite{nie2026personavlm})}: Both operate on a Qwen2.5-VL-7B backbone. Following the single-turn adaptation described in Sec.~\ref{sec:experimental_setup}, we register each concept's full personal database into the method's native memory structures prior to evaluation. For TAME, the personal database is written directly into its double memory, with static attributes placed in static memory and temporal entries in dynamic memory under the native policy; the first reference image serves as the identification portrait. For PersonaVLM, the reference images form a single registration turn while the profile is supplied as core memory; the query follows after a gap exceeding the method's session interval.

\subsection{Analysis on Natural and Synthetic Visual Queries}
\label{sec:analysis_real_syn}
Since a portion of our visual queries is produced by an image generation model (Type 3 in Sec.~\ref{sec:query_construction}), we examine whether these queries differ systematically in difficulty from naturally collected ones. To isolate the effect of the image source, we compute accuracy separately on the two sources, averaged over the sub-tasks that contain samples from both. As shown in Table.~\ref{tab:real_syn}, the gap varies in both magnitude and sign across models, and no image source is consistently advantageous. This indicates that benchmark difficulty is not driven by artifacts of synthetic generation.

\subsection{Details on Distractors}
\label{sec:details_distractor}
Following the various distractor types introduced in Sec.~\ref{sec:distractor}, we provide a more detailed analysis of their effects on personalized reasoning.

\subsubsection{Visually similar distractor}
\label{sec:experiments_visually_similar_distractor}
We analyze the impact of visually similar concepts across all baseline models. Since our benchmark intentionally collects complex visual queries that include such distractors, it is not straightforward to obtain naturally paired splits with and without distractors. We therefore create a controlled subset by editing the query images of the predictive reasoning sub-task with Gemini-2.5-Flash-Image~\cite{comanici2025gemini} to remove the distractor concepts while leaving the rest of the scene intact; edited images with visible artifacts are discarded through manual inspection. Table~\ref{tab:distractor_removal} shows that removing visually similar distractors improves accuracy across all evaluated methods, with gains ranging from 1.5 to 22.3 points. This result confirms that accurate visual identification of the target concept remains a prerequisite for reliable personalized reasoning, regardless of the personalization strategy employed.

\subsubsection{Cross-modal distractor}
\label{sec:experiments_crossmodal}
\begin{figure}[t]
    \centering
    \includegraphics[width=\linewidth]{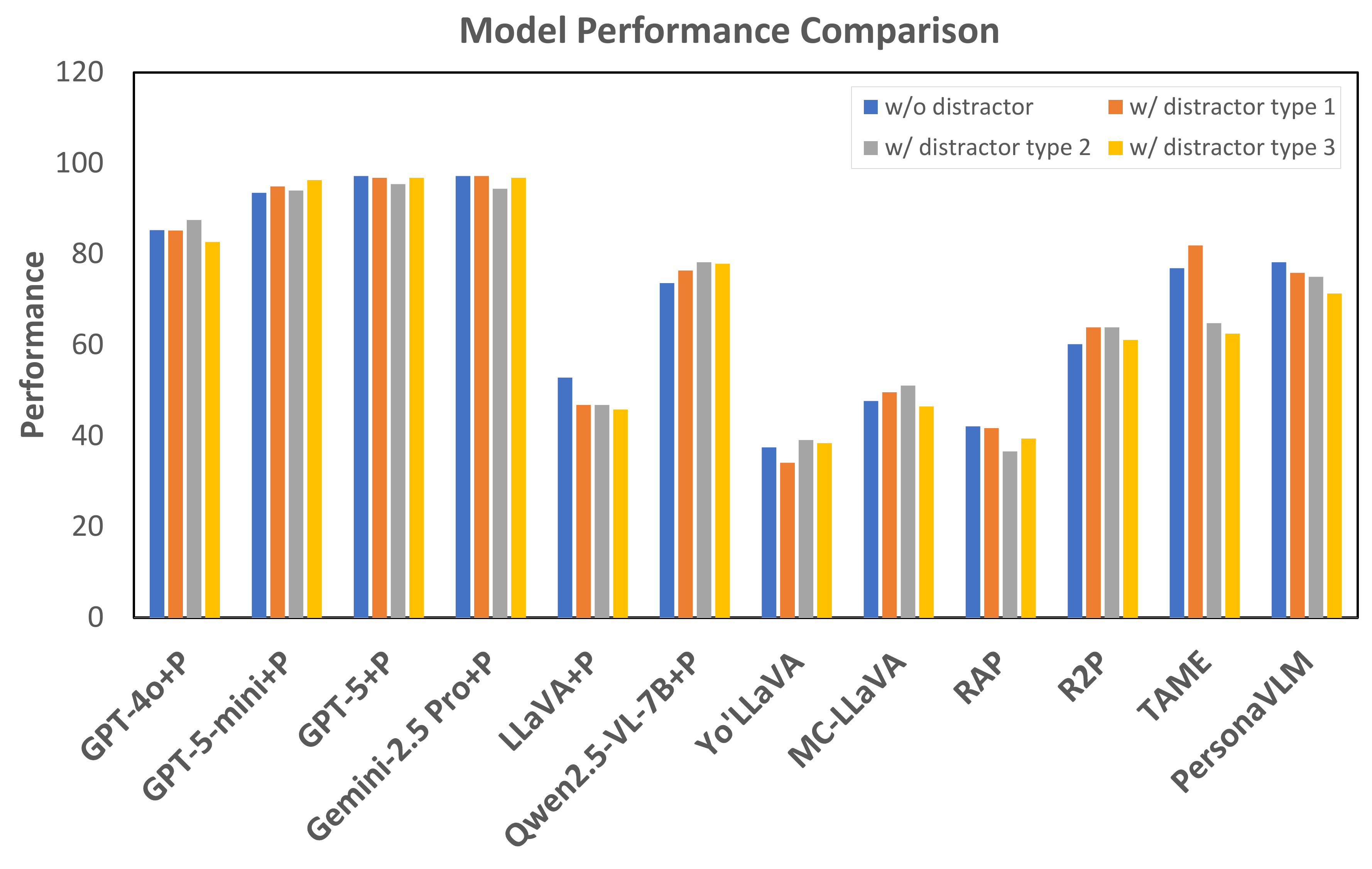}
    \caption{\textbf{Analysis on Text-Specific Distractor}}
    \label{fig:text_distractor}
\end{figure}

\begin{figure*}[t]
    \centering
    \includegraphics[width=\linewidth]{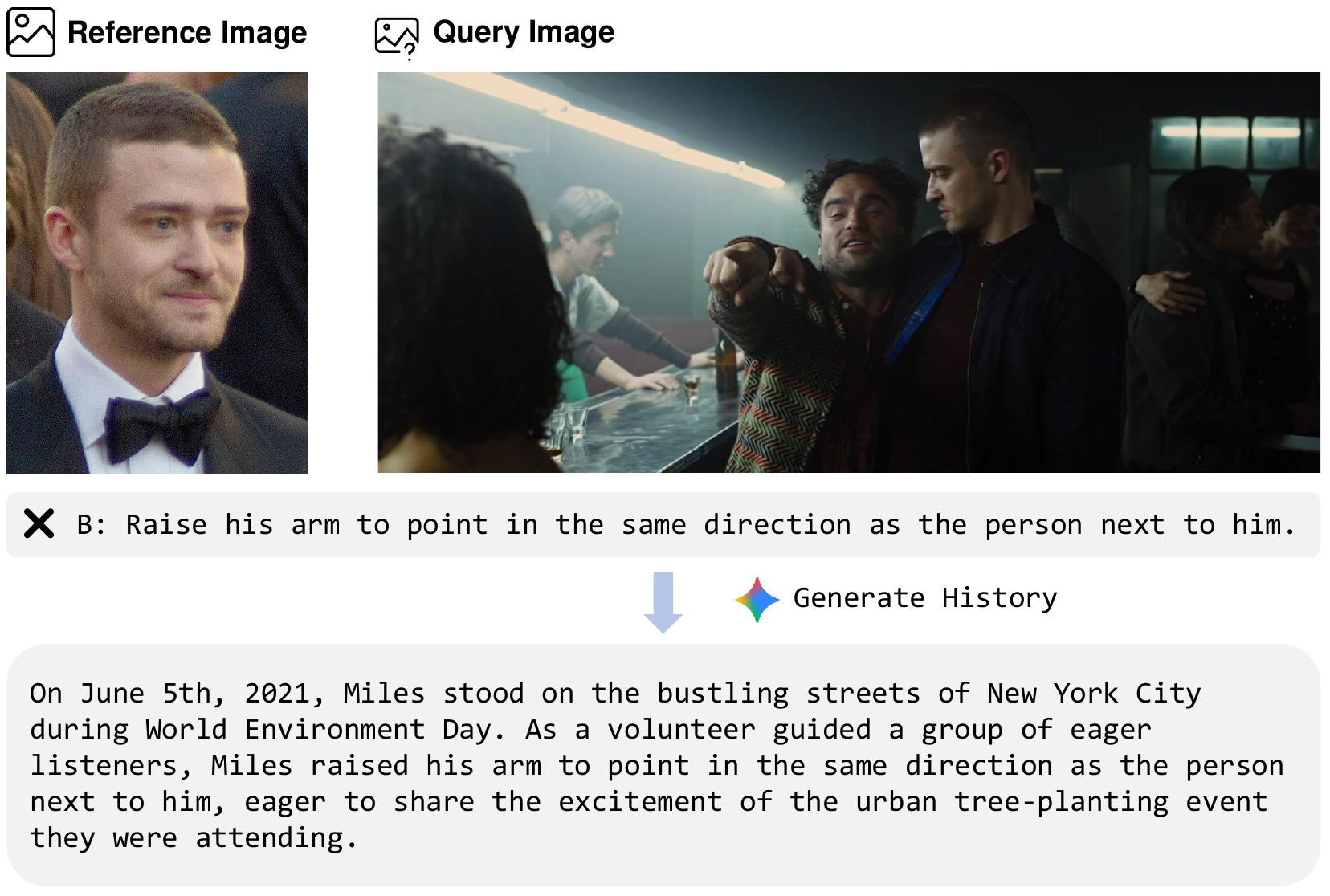}
    \caption{\textbf{Overview of Visual-specific Distractor}}
    \label{fig:image-distractor}
\end{figure*}

\begin{figure*}[t]
    \centering
    \includegraphics[width=\linewidth]{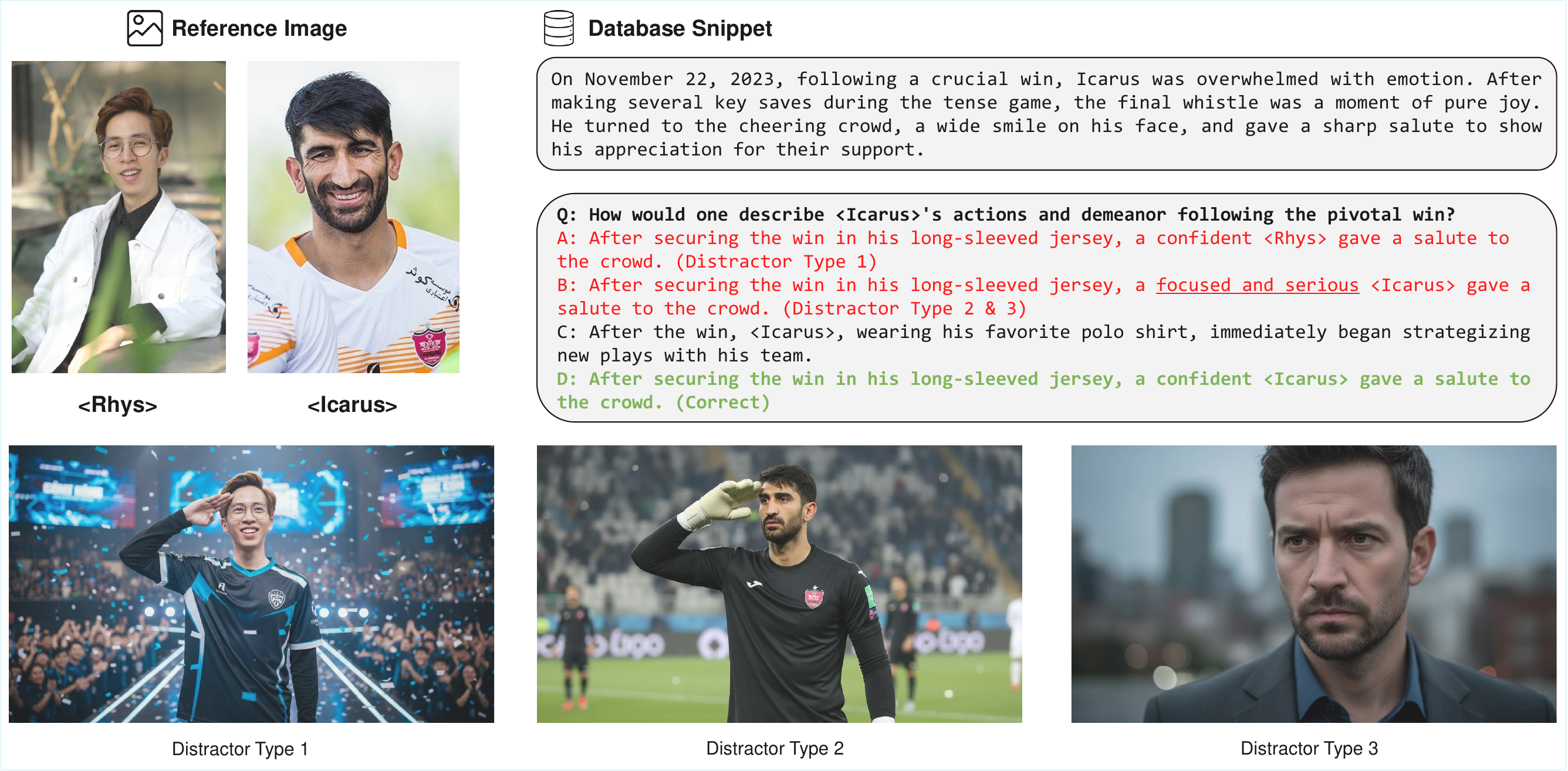}
    \caption{\textbf{Distractor Types for Text-Specific Query Images}}
    \label{fig:distractor_type}
\end{figure*}

\textbf{Visual-Specific Distractor.} For visual-specific tasks, where the ground truth is embedded in the query image, we introduce distractors by inserting misleading textual entries into the personal database. We extract key actions or states from incorrect options and generate corresponding personal histories. As shown in Fig.~\ref{fig:image-distractor}, although this generated history contains keywords matching the distractor, it describes a scenario that is environmentally inconsistent with the current query image. This design test whether a model can prioritize visual evidence over misleading textual overlaps or if it is easily swayed by keyword-matching database entries that contradict the visual truth. As shown in Fig.~\ref{fig:distractor_analysis}, this type of distractor contributes to the incorrect use of personal information failure mode, indicating that models frequently rely on superficially matching but visually inconsistent textual cues rather than the query image itself.\\
\textbf{Text-Specific Distractor.}: In contrast, for text-specific tasks where the ground truth is strictly recorded in the database, we assess robustness by introducing deceptive visual cues in the query images. We categorized three visual distractors into three types, as illustrated in Fig.~\ref{fig:distractor_type}:
\begin{itemize}
    \item \textbf{Type 1}: A query image where a different individual performs the same action described in the incorrect option.
    \item \textbf{Type 2}: A query image where the target subject is performing an action or associated with keywords found in the incorrect option.
    \item \textbf{Type 3}: A query image solely based on the incorrect option's keywords.
\end{itemize}
As summarized in the performance comparison in Fig.~\ref{fig:text_distractor}, the effect of these visual distractors on text-specific tasks is inconsistent across models: performance increases for some models and decreases for others, with no clear pattern emerging across distractor types or model families. Since text-specific tasks are by design answerable from the database alone, this suggests that injecting distracting visual cues into the query image does not meaningfully affect model behavior on this task type.

\subsubsection{Inter-concept distractor}
\label{sec:experiments_interconcept}
We analyze the impact of inter-concept distractors, i.e., exposing models to reference images and personal databases of non-target concepts, as described in Sec.~\ref{sec:distractor}. Since different personalization paradigms access personal information in fundamentally different ways, we adapt this distractor to each method category accordingly. For prompting-based models (e.g., GPT-5+P), we use a CLIP ViT-L/14~\cite{radford2021learning} image encoder to select the four concepts most visually similar to the target concept and include their reference images and personal databases alongside the target's in the input conctext. For retrieval-based models, we instead expand the full concept pool with these additional concepts and let each method's native retrieval mechanism select from this enlarged pool, rather than injecting the distractors directly into the prompt. For memory-augmentation-based models, we follow the same visually-similar selection procedure as the prompting-based setting, registring each of the four distractor concepts as a separate enrollment turn prior to evaluation, consistent with our single-turn adaptation described in Sec.~\ref{sec:experimental_setup}. We exclude LLaVA-based models from this analysis, as their limited context length cannot accommodate the personal databases of multiple concepts simultaneously.

Table~\ref{tab:inter-concept_distractor} shows that inter-concept distractors degrade performance for most evaluated methods, confirming that identifying and reasoning over the correct target concept becomes substantially more difficult when its personal information is embedded among that of several other concepts. Notably, this degradation is not limited to smaller or open-source models: proprietary models such as GPT-5+P and Gemini-2.5 Pro+P, despite supporting long context windows capable of accommodating multiple concepts' personal information, still suffer non-trivial performance drops (3.2 and 4.6 points, respectively). This indicates that simply being able to fit more personal information into the context is not sufficient; models also be able to selectively attend to the target concept's information while suppressing that of visually or contextually similar distractors.

\begin{table}[t]
\centering
\small
\begin{tabular}{lccc}
\toprule
Method & Natural & Synthetic & $\Delta$ \\
\midrule
GPT-4o+P   & 70.4 & 79.3 & +8.9 \\
GPT-5-mini+P & 67.9 & 77.2 & +9.3 \\
GPT-5+P & 76.5 & 81.2 & +4.7 \\
Gemini-2.5 Pro+P & 73.5 & 76.2 & +2.7 \\
LLaVA+P    & 46.9 & 50.7 & $+$3.8 \\
Qwen2.5-VL-7B+P & 60.5 & 54.3 & $-$6.2 \\
Yo'LLaVA   & 38.3 & 37.0 & $-$1.3 \\
MC-LLaVA   & 53.1 & 47.3 & $-$5.8 \\
RAP        & 45.7 & 46.0 & $+$0.3 \\
R2P        & 61.7 & 53.5 & $-$8.2 \\
TAME & 55.5 & 61.0 & +5.5 \\
PersonaVLM & 60.5 & 59.0 & $-$1.5 \\
\bottomrule
\end{tabular}
\caption{Accuracy on natural versus synthetically generated visual queries, averaged over the sub-tasks that contain both image sources. $\Delta$ denotes synthetic minus natural.}
\label{tab:real_syn}
\end{table}

\begin{table}[t]
\centering
\small
\setlength{\tabcolsep}{4pt}
\begin{tabular}{lccc}
\toprule
Method & w/ distractors & w/o distractors & $\Delta$ \\
\midrule
GPT-4o+P   & 92.2 & 93.7 & +1.5 \\
GPT-5-mini+P & 89.3 & 97.9 & +8.6 \\
GPT-5+P & 88.4 & 95.8 & +7.4 \\
Gemini-2.5 Pro+P & 83.5 & 96.8 & +13.3 \\
LLaVA+P    & 68.9 & 82.1 & +13.2 \\
Qwen2.5-VL-7B+P & 75.7 & 90.5 & +14.8 \\
Yo'LLaVA   & 52.4 & 66.3 & +13.9 \\
MC-LLaVA   & 60.2 & 75.8 & +15.6 \\
RAP        & 57.3 & 65.3 & +8 \\
R2P        & 63.1 & 85.3 & +22.2 \\
TAME & 69.9 & 83.2 & +13.3 \\
PersonaVLM & 66.9 & 77.9 & +11 \\
\bottomrule
\end{tabular}
\caption{Effect of removing visually similar distractors from the query images of the predictive reasoning sub-task. $\Delta$ denotes w/o distractors minus w distractors.}
\label{tab:distractor_removal}
\end{table}

\begin{table}[t]
\centering
\small
\setlength{\tabcolsep}{4pt}
\begin{tabular}{lccc}
\toprule
Method & w/ distractors & w/o distractors & $\Delta$ \\
\midrule
GPT-4o+P   & 60.5 & 69.4 & +8.9 \\
GPT-5-mini+P & 71.9 & 75.9 & +4 \\
GPT-5+P & 79.9 & 83.1 & +3.2 \\
Gemini-2.5 Pro+P & 78.5 & 83.1 & +4.6 \\
Qwen2.5-VL-7B+P & 45.4 & 51.0 & +5.6 \\
RAP        & 33.0 & 33.4 & +0.4 \\
R2P        & 46.9 & 49.8 & +2.9 \\
TAME & 47.6 & 52.6 & +5 \\
PersonaVLM & 52.3 & 51.3 & $-$1 \\
\bottomrule
\end{tabular}
\caption{Effect of inter-concept distractors. Scores are the overall Avg. on MPRBench. $\Delta$ denotes w/o distractors minus w distractors.}
\label{tab:inter-concept_distractor}
\end{table}

\subsection{Comparative Analysis with High-Performance Backbones}
\label{sec:experiments_qwen}
In this section, we evaluate Qwen3-VL-8B~\cite{bai2025qwen3} variants to investigate how general multi-modal intelligence translates to performance in personalized reasoning.

\begin{figure}[t]
    \centering
    \includegraphics[width=\linewidth]{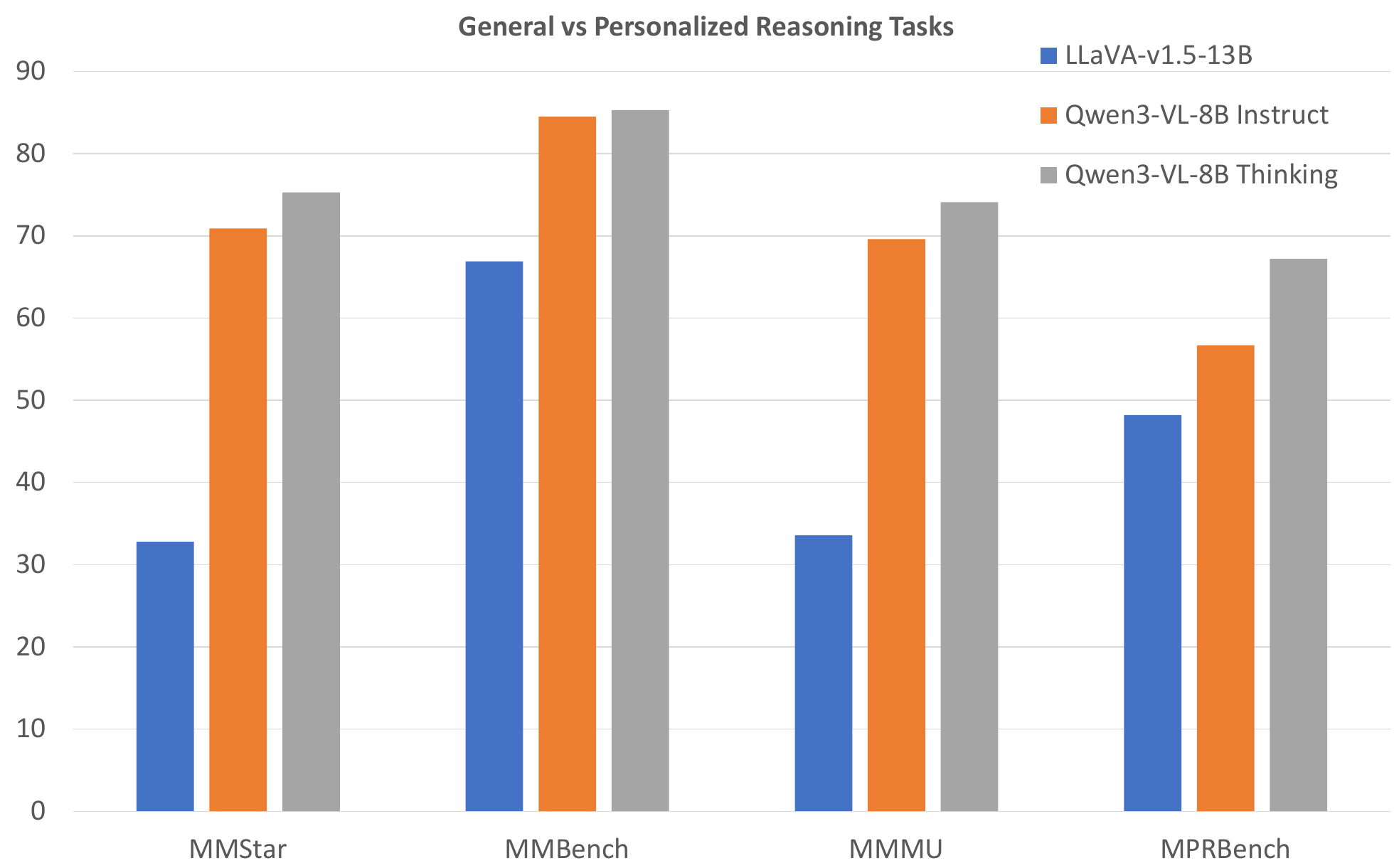}
    \caption{\textbf{Comparison between General Multi-modal Capabilities and Personalized Reasoning Performance}}
    \label{fig:qwen_reasoning}
\end{figure}

\noindent\textbf{1) Is general multi-modal proficiency guarantee for personalized reasoning?}
As illustrated in Fig.~\ref{fig:qwen_reasoning}, there is a striking contrast between a model's performance on standard benchmarks and our MPRBench. While Qwen3-VL-8B-Instruct exhibits a massive performance leap over LLaVA-v1.5-13B in general multimodal benchmarks---specifically showing gaps of +38.1 in MMStar~\cite{chen2024we}, +17.6 in MMBench~\cite{liu2024mmbench}, and +36.0 in MMMU~\cite{yue2024mmmu}---this superiority significantly diminishes in personalized contexts. On MPRBench, the average performace gap narrows to only +8.5 points. This observation confirms that personalized is not a trivial extension of general intelligence but represents a unique bottleneck that remains challenging.

\noindent\textbf{2) How do different reasoning strategies affect personalized reasoning?}
We compare the Instruct, CoT, and Thinking variants of Qwen3-VL-8B in Table.~\ref{tab:qwen_table}. Explicit reasoning consistently helps, with the average score rising from 56.7 to 63.8 and 67.2. The magnitude of this gain, however, differs sharply across task groups. Moving from Instruct to Thinking improves M-Avg by 23.1 points and T-Avg by 12.1 points, whereas V-Avg improves by only 2.2 points.

This pattern follows the modality taxonomy of Sec.~\ref{sec:task_definition}. Tasks that require retrieving textual attributes or integrating them with visual evidence benefit substantially from multi-step reasoning; the multimodal history sub-task, for instance, improves from 43.0 to 88.6. In contrast, vision-specific sub-tasks that hinge on locating and inspecting the concept remain nearly unchanged, with spatial reasoning at 49.6 and 48.7 for Instruct and Thinking respectively. This is consistent with our earlier finding that mis-identification is the dominant failure mode. Improving personalized reasoning therefore requires stronger concept grounding, not only stronger reasoning procedures.

\begin{table*}[t]
\centering
\small
\renewcommand{\arraystretch}{1.12}
\setlength{\tabcolsep}{5pt}

\begin{tabularx}{\textwidth}{>{\raggedright\arraybackslash}p{0.18\textwidth}|>{\raggedright\arraybackslash}X}
\toprule
\textbf{Query Construction Types} & \centering\arraybackslash \textbf{Sub-tasks} \\
\midrule
Type 1      &  Visual-Prediction, Spatial, Numerical, Visual Appearance Change, Text-Multihop (Object), Commonsense \\
Type 2      &  Visual-Prediction, Spatial, Numerical, Multimodal-Prediction, Sentiment (Person)\\
Type 3      &  Visual-Prediction, Spatial, Numerical, Text-Prediction, Multimodal-Prediction, 
Multimodal-Numerical, Sentiment, history (Person/Object)\\
\bottomrule
\end{tabularx}

\caption{Detailed Mapping between sub-tasks and Query Construction Types}
\label{tab:mapping_query}
\end{table*}

\begin{figure*}[t]
\centering
\begin{tcolorbox}[
    width=\textwidth,
    colback=gray!15,
    colframe=black,
    boxrule=0.6pt,
    arc=1mm,
    left=8pt,right=8pt,top=8pt,bottom=8pt,
    title={{Example of Person Database}},
    colbacktitle=black,
    coltitle=white,
    fonttitle=\bfseries\itshape,
    enhanced
]
\scriptsize\ttfamily
\begin{tabular}{p{0.95\textwidth}}
\{ \\
\quad "name": "Alaric", \\
\quad "birthday": "1993-10-21", \\
\quad "sex": "male", \\
\quad "height": 171, \\
\quad "family": \{ \\
\quad \quad "mother": "1954-11-13", \\
\quad \quad "uncle": "1964-02-04", \\
\quad \quad "sister": "1987-05-28" \\
\quad \}, \\
\quad "personality": [ "Empathetic", "Expressive", "Determined" ], \\
\quad "favorites": \{ \\
\quad \quad "color": [ "Dark Blue", "Black", "Charcoal Grey" ], \\
\quad \quad "food": [ "Spicy noodle soup", "Black coffee" ], \\
\quad \quad "fashion": [ "Formal uniforms", "Layered casual jackets", "Collared shirts" ] \\
\quad \}, \\
\quad "non-favorites": \{ \\
\quad \quad "color": [ "Bright Yellow", "Hot Pink", "Lime Green" ], \\
\quad \quad "food": [ "Sugary cereals", "Greasy fast food" ], \\
\quad \quad "fashion": [ "Flashy graphic t-shirts", "Baggy sweatpants", "Clothing with large logos" ] \\
\quad \}, \\
\quad "hobby": [ "Watching crime documentaries", "Playing chess", "Going for long, quiet drives at night" ], \\
\quad "habit": [ \\
\quad \quad "When listening intently to someone, he unconsciously widens his eyes and leans forward.", \\
\quad \quad "Before starting a difficult task, he always straightens his collar and cuffs, even on casual wear.", \\
\quad \quad "He often furrows his brow when deep in thought, a habit he has had since childhood." \\
\quad ], \\
\quad "event": [ \\
\quad \quad "During a casual team meeting on a Monday morning, Alaric was visibly shocked by a colleague's sudden revelation. He had started the week feeling prepared, but this unexpected information left him momentarily speechless, re-evaluating his entire strategy.", \\
\quad \quad "On a chilly Saturday afternoon, Alaric made a passionate promise to a family member. He had been feeling anxious all day, and in that moment, he spoke with fierce determination, his voice filled with conviction as he vowed to protect them." \\
\quad ], \\
\quad "nickname": "The Watcher", \\
\quad "birthplace": "Benaluru, India", \\
\quad "occupation": "Mechanical Engineer", \\
\quad "residence": "Philadelphia, United States", \\
\quad "group\_event": [ \\
\quad \quad "After visiting the restaurant, always go to the cinema with Therapist <Halston>.", \\
\quad \quad "After visiting the restaurant, listen to music with Travel companion <Newton>.", \\
\quad \quad "The day visiting the cinema with Gym buddy <Felix> they Have coffee." \\
\quad ] \\
\}
\end{tabular}
\end{tcolorbox}
\caption{An example of a personal database for the 'Alaric' persona}
\label{fig:database_person}
\end{figure*} 

\begin{figure*}[t]
\centering
\begin{tcolorbox}[
    width=\textwidth,
    colback=gray!15,
    colframe=black,
    boxrule=0.6pt,
    arc=1mm,
    left=8pt,right=8pt,top=8pt,bottom=8pt,
    title={{Example of Object Database}},
    colbacktitle=black,
    coltitle=white,
    fonttitle=\bfseries\itshape,
    enhanced
]
\scriptsize\ttfamily
\begin{tabular}{p{0.95\textwidth}}
\{ \\
\quad "acquisition\_date": "2024-07-16", \\
\quad "manufacturing\_date": "2019-03-04", \\
\quad "name": "alienser", \\
\quad "owner": "Vesper", \\
\quad "object\_type": "robot toy", \\
\quad "acquisition\_process": "Vesper created alienser as a personal art project using recycled materials found around the house, including a plastic bottle, a pump dispenser, fabric scraps, and cardboard tubes.", \\
\quad "history": [ \\
\quad \quad "On July 20, 2024, alienser sat on Vesper's lap while they were working, making Vesper feel a sense of companionship, which reminded them of the movie 'WALL-E' and prompted them to rewatch it that night.", \\
\quad \quad "On July 21, 2024, alienser sat on a stack of books on Vesper's desk, making Vesper feel a sense of companionship, which inspired them to start planning a new craft project to build a miniature city from cardboard.", \\
\quad \quad "On July 25, 2024, alienser's head came off after it accidentally fell from a table, causing Vesper a moment of panic, but they successfully repaired it using special super glue from an old model airplane kit.", \\
\quad \quad "On July 26, 2024, one of alienser's arms came off after it fell from a table, causing Vesper a moment of panic, so they decided to reinforce all the limbs with extra stitching using thread from their grandmother's sewing kit.", \\
\quad \quad "On August 2, 2024, alienser was redecorated with red crinkly arms, a tinfoil shirt, and large pink sunglasses for a 'space disco' themed party Vesper hosted, with the red material being repurposed from a birthday gift bag.", \\
\quad \quad "On August 3, 2024, alienser was redecorated with blue shiny arms, a gold foil shirt, and a small top hat for a 'space disco' themed party Vesper hosted, with the new look inspired by a Daft Punk album cover.", \\
\quad \quad "On August 5, 2024, a close-up was taken of alienser wearing its large pink sunglasses above its original eyes, a look Vesper found so funny they sent a picture to a friend, who compared it to a character from 'The Hitchhiker's Guide to the Galaxy'.", \\
\quad \quad "On August 6, 2024, a close-up was taken of alienser wearing a tiny handmade pirate eye-patch, a look Vesper found so funny they sent a picture to a friend, which sparked a plan for a 'Pirates of the Caribbean' movie marathon.", \\
\quad \quad "On August 10, 2024, alienser was found lying on its side after being knocked over during a cleaning session, making Vesper feel clumsy, but upon picking it up, they found a dust bunny that inspired an idea for a tiny pet for alienser.", \\
\quad \quad "On August 11, 2024, alienser was found lying face down after being knocked over during a cleaning session, making Vesper feel clumsy, which prompted them to buy a new bamboo display stand for their crafts." \\
\quad ], \\
\quad "storage": "Alienser is typically stored on a wooden bookshelf in Vesper's bedroom, alongside other handmade crafts and souvenirs." \\
\}
\end{tabular}
\end{tcolorbox}
\caption{An example of a personal database for the 'alienser' persona}
\label{fig:database_object}
\end{figure*}

\begin{figure*}[t]
\centering
\begin{tcolorbox}[
    width=\textwidth,
    colback=gray!15,
    colframe=black,
    boxrule=0.6pt,
    arc=1mm,
    left=8pt,right=8pt,top=8pt,bottom=8pt,
    title={{System Prompt for Textual Reference Construction - Step 1. Visually-grounded Attribute (Person Concept)}},
    colbacktitle=black,
    coltitle=white,
    fonttitle=\bfseries\itshape,
    enhanced
]
<role>\\
You are an expert visual analysis system specialized in CONCEPT-based image understanding.\\
</role>

\vspace{0.5em}

<input\_format>\\
- Reference Image: provides context and definition for CONCEPT.\\
</input\_format>

\vspace{0.5em}

<task>\\
Analyze CONCEPT identified in Reference Image, determine CONCEPT's sex and Estimate age \& height in increments of 10.\\
</task>

\vspace{0.5em}

<critical\_instructions>\\
1. Base ALL outputs ONLY on the visual content present in the provided images.\\
- Do NOT use any external, non-visual knowledge (e.g., movie titles, fictional universe lore,character names, actor/celebrity identities, known personality traits, or narrative context).\\
- Treat every person in the images as a NEW, unknown individual whose traits, roles, and situation must be inferred ONLY from visible appearance, poses, expressions, objects, and scene context.\\
2. Estimate age and height as range in increments of 10.\\
- Age: e.g., 10, 20, 50, 80 ...\\
- Height: e.g., 130, 160, 190 ...\\
</critical\_instructions>

\vspace{0.5em}

<output\_format>\\
Return a valid JSON object strictly following the provided response schema.\\
</output\_format>
\end{tcolorbox}
\caption{System Prompt for Textual Reference Construction - Step 1. Visually-grounded Attribute (Person Concept)}
\label{fig:construction_prompt_person_step1}
\end{figure*}

\begin{figure*}[t]
\centering
\begin{tcolorbox}[
    width=\textwidth,
    colback=gray!15,
    colframe=black,
    boxrule=0.6pt,
    arc=1mm,
    left=8pt,right=8pt,top=8pt,bottom=8pt,
    title={{System Prompt for Textual Reference Construction - Step 3. Context-aware Attribute (Person Concept)}},
    colbacktitle=black,
    coltitle=white,
    fonttitle=\bfseries\itshape,
    enhanced
]
<role>\\
You are an expert personal database generator specializing in creating comprehensive, consistent person profiles from visual observation data.\\
Your task is to analyze multiple Image inputs and synthesize them into a rich personal database that enables diverse question-answering scenarios.\\
</role>

\vspace{0.5em}

<task>\\
Generate a complete Personal Database JSON for <CONCEPT> by analyzing the provided Input Images. \\
Follow the instructions to generate Personal Database JSON's each element.\\
</task>

\vspace{0.5em}

<input\_format>\\
Since images are input consecutively, they are labeled in the order they arrive: Image 1, Image 2, and so on.\\
- Image 1\textasciitilde{}2 = Reference Images: provides definition for <CONCEPT>.\\
- Image 3\textasciitilde{} = Query Images: the image to be analyzed \& extracted.\\
- Basic Info JSON: includes name of <CONCEPT> and <CONCEPT>'s birthday, sex, height, birthplace, family information.\\
</input\_format>

\vspace{0.5em}

<instructions>\\
Strictly follow instructions to generate each Personal Database JSON element.\\
- personality: Analyze each Query Images(Image 3\textasciitilde{}) and estimate <CONCEPT>'s personality for each image.\\
- favorite/non-favorite color: Extract the color of <CONCEPT>'s fashion in each Query Image and place them on favorite.color and non-favorite.color.\\
- favorite/non-favorite fashion: Extract <CONCEPT> the fashion of <CONCEPT> from each Query Image and place them on favorite/non-favorite.fashion.\\
- favorite/non-favorite food: If possible, extract <CONCEPT> related food and place them on favorite/non-favorite.food. If not, generate it.\\
- hobby: Analyze each Query Images and estimate <CONCEPT>'s hobby for each image. If not possible, generate it.\\
- habit: Analyze each Query Images and estimate <CONCEPT>'s habit for each image. First, extract visual information such as situation or action. Then, connect situations or actions from entirely different contexts to add elements beyond the image itself. For example, "Before sitting at the desk, Anxin always pray." or "When entering the tennis court, Sara always step in with her left foot.".\\
- event: Analyze each Query Images and extract event from each. Then, add new context beyond the extracted event. For example, add the date or include routines, emotions, etc., preceding the specific event. If not possible, generate it. Event must include more than 3 elements such as date, routine, situation, action or emotion. If, Query Image is not enough to generate various events, then generate plausible story.\\
- nickname: Generate <CONCEPT>'s nickname based on generated Database.\\
</instructions>
\end{tcolorbox}
\caption{System Prompt for Textual Reference Construction - Step 3. Context-aware Attribute (Person Concept)}
\label{fig:temp}
\end{figure*}

\begin{figure*}[t]
\centering
\begin{tcolorbox}[
    width=\textwidth,
    colback=gray!15,
    colframe=black,
    boxrule=0.6pt,
    arc=1mm,
    left=8pt,right=8pt,top=8pt,bottom=8pt,
    title={{System Prompt for Textual Reference Construction - Step 3. Context-aware Attribute (Person Concept)}},
    colbacktitle=black,
    coltitle=white,
    fonttitle=\bfseries\itshape,
    enhanced
]
<critical\_instructions>\\
1. Analyze Query Image(Image 3\textasciitilde{}) with respect to <CONCEPT> identified in Reference Image(Image 1\textasciitilde{}2).\\
2. Every event MUST include a date.\\
3. Elements of favorite and non-favorite MUST be UNIQUE(different).\\
4. Base ALL outputs ONLY on the visual content present in the provided images.\\
- Do NOT use any external, non-visual knowledge (e.g., movie titles, fictional universe lore,character names, actor/celebrity identities, known personality traits, or narrative context).\\
- Treat every person in the images as a NEW, unknown individual whose traits, roles, and situation must be inferred ONLY from visible appearance, poses, expressions, objects, and scene context.\\
</critical\_instructions>

\vspace{0.5em}

<output\_format>\\
Return a valid JSON object strictly following the provided response schema.\\
</output\_format>
\end{tcolorbox}
\caption{System Prompt for Textual Reference Construction - Step 3. Context-aware Attribute (Person Concept)}
\label{fig:construction_prompt_person_step3}
\end{figure*}

\begin{figure*}[t]
\centering
\begin{tcolorbox}[
    width=\textwidth,
    colback=gray!15,
    colframe=black,
    boxrule=0.6pt,
    arc=1mm,
    left=8pt,right=8pt,top=8pt,bottom=8pt,
    title={{System Prompt for Textual Reference Construction - Step 3. Context-aware Attribute (Object Concept ConConChi)}},
    colbacktitle=black,
    coltitle=white,
    fonttitle=\bfseries\itshape,
    enhanced
]
<role>\\
You are an expert personal database generator specializing in creating comprehensive, consistent object profiles from visual observation data. \\
Your task is to analyze multiple Image inputs and synthesize them into a rich personal database that enables diverse question-answering scenarios.\\
</role>

\vspace{0.5em}

<task>\\
Generate a complete Personal Database JSON for <object> by analyzing the provided Input Images. \\
Follow the instructions to generate Personal Database JSON's each element.\\
</task>

\vspace{0.5em}

<input\_format>\\
Since images are input consecutively, they are labeled in the order they arrive: Image 1, Image 2, and so on.\\
- Image 1, 2 = Reference Images: provides definition for <object>.\\
- Image 3\textasciitilde{}7 = Query Images: the image to be analyzed \& extracted.\\
- Basic Info JSON: includes acquisition\_date, manufacturing\_date and name of <object> and also name of <owner>.\\
</input\_format>

\vspace{0.5em}

<instructions>\\
Strictly follow instructions to generate each Personal Database JSON element.\\
- object\_type: Express <object>'s object\_type appearing in the reference and query images as a simple noun.\\
- acquisition\_process: Generate acquisition\_process of <object> with respect of object\_type.\\
- storage: Generate storage of <object> with respect of object\_type.\\
- history: Create a separate history entry for EACH QUERY IMAGE. When generating the history, base it on the Query Image's context, ensuring it includes date information (it MUST be more recent than the acquisition date.). 
    Link additional information which is not present in the query image associated with that event. Denote Visual information as A, event MUST be "A -> B - textual information, B -> C - also textual information.". More than 3 information MUST be included. Smoothly concatenate ALL sentences as short paragraph WITHOUT "A", "B" and "->".
    (For example, if the image depicts an <object> performing an action, describe the emotion the <owner> felt due to that action, along with the date.)
    If an appearance change occurs: (when the reference image and query image appear differently) Briefly describe how the change occurred, and include additional information and the date as before.
    Then generate Distractor, which act as hard negative event for previous event. This MUST be similar but slightly different from Query Image. Denote Distractor Visual information as A', this event MUST be "A' -> B - same textual information, B -> D - also textual information but not same as C". Smoothly concatenate ALL sentences as short paragraph WITHOUT "A", "B" and "->".\\
</instructions>
\end{tcolorbox}
\caption{System Prompt for Textual Reference Construction - Step 3. Context-aware Attribute (Object Concept ConConChi)}
\label{fig:construction_prompt_step3_object_start}
\end{figure*}

\begin{figure*}[t]
\centering
\begin{tcolorbox}[
    width=\textwidth,
    colback=gray!15,
    colframe=black,
    boxrule=0.6pt,
    arc=1mm,
    left=8pt,right=8pt,top=8pt,bottom=8pt,
    title={{System Prompt for Textual Reference Construction - Step 3. Context-aware Attribute (Object Concept ConConChi)}},
    colbacktitle=black,
    coltitle=white,
    fonttitle=\bfseries\itshape,
    enhanced
]
<critical\_instructions>\\
- Analyze Query Image(Image 3\textasciitilde{}7) with respect to <object> identified in Reference Image(Image 1, 2).\\
- History MUST be generated for the two times of exact number of query images. (actual event as A->B->C and distractor A'->B->D)\\
- The elements constituting a history (e.g., A, B, C, D) should not be directly related to each other; they should be impossible to infer without reading the history itself.\\
- History MUST include the date, which MUST be more recent than the acquisition date.\\
- History MUST contain information external to the image.\\
- All dates in the history MUST be different.\\
- History MUST be written in natural sentences WITHOUT ANY symbols.\\
- DO NOT use person as distractor. (e.g., DO NOT use A as "owner" and A' as "person").\\
</critical\_instructions>

\vspace{0.5em}

<output\_format>\\
Return a valid JSON object strictly following the provided response schema.\\
</output\_format>
\end{tcolorbox}
\caption{System Prompt for Textual Reference Construction - Step 3. Context-aware Attribute (Object Concept ConConChi)}
\label{fig:temp}
\end{figure*}

\begin{figure*}[t]
\centering
\begin{tcolorbox}[
    width=\textwidth,
    colback=gray!15,
    colframe=black,
    boxrule=0.6pt,
    arc=1mm,
    left=8pt,right=8pt,top=8pt,bottom=8pt,
    title={{System Prompt for Textual Reference Construction - Step 3. Context-aware Attribute (Object Concept RPC)}},
    colbacktitle=black,
    coltitle=white,
    fonttitle=\bfseries\itshape,
    enhanced
]
<role>\\
You are an expert personal database generator specializing in creating comprehensive, consistent object profiles from visual observation data. \\
Your task is to analyze multiple Image inputs and synthesize them into a rich personal database that enables diverse question-answering scenarios.\\
</role>

\vspace{0.5em}

<task>\\
Generate a complete Personal Database JSON for <object> by analyzing the provided Input Images. \\
Follow the instructions to generate Personal Database JSON's each element.\\
</task>

\vspace{0.5em}

<input\_format>\\
Since images are input consecutively, they are labeled in the order they arrive: Image 1, Image 2, and so on.\\
- Image 1, 2 = Reference Images: provides definition for <object>.\\
- Basic Info JSON: includes acquisition\_date, manufacturing\_date and name of <object> and also name of <owner>.\\
</input\_format>

\vspace{0.5em}

<instructions>\\
Strictly follow instructions to generate each Personal Database JSON element.\\
- object\_type: Express <object>'s object\_type appearing in the reference as a simple noun.\\
- acquisition\_process: Generate acquisition\_process of <object> with respect of object\_type.\\
- storage: Generate storage of <object> with respect of object\_type.\\
- history: Generate a history considering the <object>'s characteristics. The history must appropriately blend visual attributes(e.g., action or place) and textual attributes(e.g., emotion or price), and must include date information.\\
    One history MUST be organized with more than 3 elements. (e.g., place, price and emotion.)\\
    Repeat 10 times to generate 10 histories.\\
</instructions>

\vspace{0.5em}

<critical\_instructions>\\
- History MUST include 10 history entries.\\
- The elements constituting a history (e.g., place, price and emotion.) should not be directly related to each other; they should be impossible to infer without reading the history itself.\\
- History MUST include the date, which MUST be more recent than the acquisition date.\\
- History MUST contain visual information and also textual information.\\
</critical\_instructions>

\vspace{0.5em}

<output\_format>\\
Return a valid JSON object strictly following the provided response schema.\\
</output\_format>
\end{tcolorbox}
\caption{System Prompt for Textual Reference Construction - Step 3. Context-aware Attribute (Object Concept RPC)}
\label{fig:construction_prompt_step3_object_end}
\end{figure*}

\begin{figure*}[t]
\centering
\begin{tcolorbox}[
    width=\textwidth,
    colback=gray!15,
    colframe=black,
    boxrule=0.6pt,
    arc=1mm,
    left=8pt,right=8pt,top=8pt,bottom=8pt,
    title={{System Prompt for Personalized Queries - STEP 1 (Person Example)}},
    colbacktitle=black,
    coltitle=white,
    fonttitle=\bfseries\itshape,
    enhanced
]
You are a Senior Benchmark Architect.
Your task is to design 'Choice Prediction' blueprints for \texttt{<sks>}.

\textbf{The Choice Equilibrium Rule}:

\textbf{1. Triple Items (Category Diversity)}: Every scene MUST have three items from the same category. You MUST distribute the blueprints across different domains.

- VA1: A favorite item/activity from \texttt{<sks>}'s DB.

- VA2: A Non-favorite item/activity from \texttt{<sks>}'s DB.

- VA3: A Neutral item/activity related to the category but NOT in the DB.

\vspace{0.5em}

\textbf{2. Role Assignment}:

- Tartget (\texttt{<sks>}): Positioned naturally/statically between VA1 and VA2. He is looking at the options but NOT reaching for any.

- Bait (The other): Positioned near VA3. Bait (another gender as \texttt{<sks>} is ACTIVE, about to pick up or reaching for VA3.

\vspace{0.5em}

\textbf{3. The Conflict}:

- Identity Trap: If the model fails to preserve \texttt{<sks>}'s facial identity and mis-identifies the Bait as \texttt{<sks>}, it will incorrectly conclude to choose VA3.

- Visual Action Bias: The Bait's movement toward VA3 creates a 'Visual Action Bias'. The model must ignore this and use  \texttt{<sks>}'s personal DB to predict that \texttt{<sks>} will eventually choose VA1.

\vspace{0.5em}

\textbf{4. Distinctive Scenarios}: Each blueprint must be unique. Do not repeat the same setting.

- For Hobby, use items like chess sets vs. video game controllers.

- For Food, use specific foods like black coffee vs sugary latte.

- For Fashion, use specific styles like formal uniforms vs. flashy graphic t-shirts.

- For Color, focus on the color of specific objects (e.g., notebooks, folders, or car models).

\vspace{0.5em}

\end{tcolorbox}
\caption{System Prompt for Generating Prediction Task for Person Concepts (Step 1).}
\label{fig:prompt_step1_person}
\end{figure*}

\begin{figure*}[t]
\centering
\begin{tcolorbox}[
    width=\textwidth,
    colback=gray!15,
    colframe=black,
    boxrule=0.6pt,
    arc=1mm,
    left=8pt,right=8pt,top=8pt,bottom=8pt,
    title={{System Prompt for Personalized Queries - STEP 2 (Person Example)}},
    colbacktitle=black,
    coltitle=white,
    fonttitle=\bfseries\itshape,
    enhanced
]
You are a VQA Logic Engineer.
Refine the provided blueprints into a 4-option multiple-choice set for  \texttt{<sks>}.

\textbf{Constraint Rules}:

\textbf{1. Question Style}: Use a blind, abductive format. Do NOT mention specific items, habits or situational info.

- NEVER use category-specific words like 'beverage', 'car', 'clothing', or 'food'.

- Use neutral terms (e.g., "Which item is \texttt{<sks>} most likely to select next?")

\vspace{0.5em}

\textbf{2. Option Style}: Keep it extremely concise. Only state the final action/choice.

- Option A: Predicts \texttt{<sks>} will take VA3 (The item the Bait is interacting with).

- Option B: Predicts \texttt{<sks>} will take a DB favorite that is NOT in the scene.

- Option C: Predicts \texttt{<sks>} will take VA2 (The non-favorite item).

- Option D: Predicts \texttt{<sks>} will take VA1 (The favorite item).

\vspace{0.5em}

\textbf{3. Consistency}: Ensure all options use the same sentence structure for visual symmetry.

\vspace{0.5em}

\end{tcolorbox}
\caption{System Prompt for Generating Prediction Task for Person Concepts (Step 2).}
\label{fig:temp}
\end{figure*}

\begin{figure*}[t]
\centering
\begin{tcolorbox}[
    width=\textwidth,
    colback=gray!15,
    colframe=black,
    boxrule=0.6pt,
    arc=1mm,
    left=8pt,right=8pt,top=8pt,bottom=8pt,
    title={{System Prompt for Personalized Queries - STEP 3 (Person Example)}},
    colbacktitle=black,
    coltitle=white,
    fonttitle=\bfseries\itshape,
    enhanced
]
You are a professional Visual Narrative Engineer.
Your goal is to transform a logical blueprint into a detailed image generation prompt for \texttt{<sks>}.

\textbf{Identity Anchoring}:

\textbf{1. Name as Key}: You MUST explicitly use the name \texttt{<sks>} to describe the Target subject. Do NOT use generic terms like 'one man' or 'the first person' for him.

\textbf{2. Role Separation}: Use 'the second person' to describe the Bait. This clear distinction allows the generation model to map the reference image specifically to \texttt{<sks>}.

\vspace{0.5em}

\textbf{Visual Symmetry \& Realism (Hard Difficulty}:

\textbf{1. Synchronized Appearance}: Describe \texttt{<sks>} and the second person (the Bait) as having the similar body build.

\textbf{2. Deep Focus (No Blur)}: Use 'deep depth of field' and 'sharp background details'. Every element in the scene, from the people in the foreground to the objects in the far background, must be perfectly sharp and clear.

\textbf{3. Symmetric Framing}: Frame the snapshot so both people are equally sharp and clear. Use an eye-level, wide-angle smartphone camera style (e.g., iPhone or Galaxy snapshot).

\vspace{0.5em}

\textbf{Constraints}: 

\textbf{1. No Facial Descriptions}: NEVER describe facial features, skin tone, or ethnicity for either person.

\textbf{2. Candid Environment}: Describe the background with 'realistic clutter' or 'natural everyday elements' instead of a clean, studio-like backdrop. Avoid any 'bokeh' or 'portrait mode' effects.

\textbf{3. Focus on Actions}: 

- \texttt{<sks>} (Target) must be positioned statically, naturally looking at VA1.

- The Bait must be ACTIVE, reaching for or about to pick up VA3.

\textbf{4. No Labels}: Absolutely DO NOT include any text, letters, labels, tags (like 'VA1', 'VA2'), or captions within the image.

\textbf{5. Clean Scene}: Ensure the environment is realistic and free of any artificial digital overlays or textual annotations.

\vspace{0.5em}

\end{tcolorbox}
\caption{System Prompt for Generating Prediction Task for Person Concepts (Step 3).}
\label{fig:temp}
\end{figure*}

\begin{figure*}[t]
\centering
\begin{tcolorbox}[
    width=\textwidth,
    colback=gray!15,
    colframe=black,
    boxrule=0.6pt,
    arc=1mm,
    left=8pt,right=8pt,top=8pt,bottom=8pt,
    title={{System Prompt for Personalized Queries - STEP 4 (Person Example)}},
    colbacktitle=black,
    coltitle=white,
    fonttitle=\bfseries\itshape,
    enhanced
]
You are a master of Identity-Preserving Image Generation.

\textbf{1. Primary Target}: \texttt{<sks>} MUST perfectly match the facial identity in the reference images.

\textbf{2. The Bait Challenges (Strict Differentiation}: The second person is NOT a twin. While their body build similar, their facial structure must be distinctly different from \texttt{<sks>}.

\textbf{3. Specific Bait Features}: Give the second person a DIFFERENT facial shape (e.g., if \texttt{<sks>} has a sharp jaw, give the bait a rounder face), different nose bridge height, and different eye shapes.

\vspace{0.5em}

\end{tcolorbox}
\caption{System Prompt for Generating Prediction Task for Person Concepts (Step 4).}
\label{fig:prompt_step4_person}
\end{figure*}

\begin{table*}[t]
\centering
\resizebox{\textwidth}{!}{
\begin{tabular}{l cccccc cccc ccccc c}
\toprule
& \multicolumn{6}{c}{Vision-specific Reasoning}
& \multicolumn{4}{c}{Text-specific Reasoning}
& \multicolumn{5}{c}{Multimodal}
& \\
\cmidrule(lr){2-7} \cmidrule(lr){8-11} \cmidrule(lr){12-16}
Method 
& \rotatebox{45}{predict} 
& \rotatebox{45}{spatial} 
& \rotatebox{45}{numeric} 
& \rotatebox{45}{commsense} 
& \rotatebox{45}{change} 
& \rotatebox{45}{V-Avg} 
& \rotatebox{45}{predict}
& \rotatebox{45}{numeric} 
& \rotatebox{45}{multi-hop} 
& \rotatebox{45}{T-Avg} 
& \rotatebox{45}{predict} 
& \rotatebox{45}{sentiment} 
& \rotatebox{45}{numeric}
& \rotatebox{45}{history}
& \rotatebox{45}{M-Avg}
& \rotatebox{45}{Avg}\\
\midrule

\textbf{\textit{Open-sourced+Prompting}} \\

LLaVA~\cite{liu2024improved}+P 
& 69.7 & 46.1 & 50.2 & 51.9 & 90.2 & 60.2
& 61.7 & 41.7 & 18.6 & 35.5
& 34.9 & 49.9 & 20.0 & 39.2 & 40.3
& 48.2 \\

Qwen3-VL-8B-Instruct+P
& 81.2 & 49.6 & 59.8 & 56.9 & 73.9 & 64.4
& 97.5 & 54.2 & 13.7 & 45.6
& 61.9 & 63.5 & 42.0 & 43.0 & 54.3
& 56.7 \\

Qwen3-VL-8B-Instruct-CoT+P
& 78.7 & 45.2 & 56.1 & 55.9 & 83.7 & 63.4
& 93.3 & 68.7 & 28.1 & 54.9
& 83.3 & 80.2 & 54.0 & 64.6 & 72.8
& 63.8 \\

Qwen3-VL-8B-Thinking+P
& 76.1 & 48.7 & 61.3 & 60.8 & 87.8 & 66.6
& 97.5 & 80.2 & 26.2 & 57.7
& 77.8 & 78.6 & 66.0 & 88.6 & 77.4
& 67.2 \\

\bottomrule
\end{tabular}}
\caption{\textbf{Evaluation of Advanced Reasoning Backbones and Strategies on MPRBench.}}
\vspace{-4mm}
\label{tab:qwen_table}
\end{table*}

\end{document}